\documentclass[twoside,11pt]{article}

\usepackage{jmlr2e}

\usepackage{amsmath,amssymb,xspace}

\usepackage[utf8]{inputenc} 
\usepackage[T1]{fontenc}    
\usepackage{hyperref}       
\usepackage{url}            
\usepackage{booktabs}       
\usepackage{amsfonts}       
\usepackage{nicefrac}       
\usepackage{microtype}      
\usepackage{xcolor}         

\usepackage{microtype}
\usepackage{graphicx}
\usepackage{booktabs} 

\usepackage{microtype}
\usepackage{graphicx}

\usepackage{natbib}

\usepackage{algorithm}
\usepackage{algorithmic}
\usepackage{paralist}
\usepackage{multirow}
\usepackage{times}
\usepackage{wrapfig}

\usepackage{url,enumerate}
\usepackage{color,xcolor}
\usepackage{makeidx}  
\usepackage{mathtools}
\usepackage{xspace}
\usepackage{epstopdf}
\usepackage{cite}

\usepackage{mathrsfs}
\usepackage{times}
\usepackage{enumerate}
\usepackage{color}
\usepackage{graphicx,epsfig}
\usepackage{url}
\usepackage{hyperref}
\usepackage{bm}
\usepackage{bbm}
\usepackage{upgreek}
\usepackage{cleveref}
\usepackage{multirow}
\usepackage{ulem}
\usepackage{cancel}
\usepackage{subcaption}
\usepackage{dsfont}
\usepackage{adjustbox}

\renewcommand{\d}{{\rm d}}  

\newcommand{\f}{{\bf f}}
\newcommand{\g}{{\bf g}}

\renewcommand{\u}{{\bf u}}
\renewcommand{\v}{{\bf v}}

\newcommand{\x}{{\bf x}}

\newcommand{\D}{{\bf D}}

\renewcommand{\H}{{\bf H}}
\newcommand{\I}{{\bf I}}

\newcommand{\Acal}{\mathcal{A}}

\newcommand{\Dcal}{\mathcal{D}}
\newcommand{\Ocal}{\mathcal{O}}

\newcommand{\Lcal}{\mathcal{L}}
\newcommand{\Tcal}{\mathcal{T}}

\newcommand{\Scal}{\mathcal{S}}

\newcommand{\Mcal}{{\mathcal{M}}}

\newcommand{\bpsi}{\boldsymbol{\psi}}

\newcommand{\blambda}{\boldsymbol{\lambda}}

\newcommand{\btheta}{\boldsymbol{\theta}}

\newcommand{\0}{{\bf 0}}

\newcommand{\ben}{\begin{enumerate}}
\newcommand{\een}{\end{enumerate}}

\newcommand{\argmin}{\operatornamewithlimits{argmin}}

\newcommand{\whJ}{{\widehat{J}}}

\newcommand{\ours}{{A-MAML}\xspace}
\newcommand{\bmaml}{{{B-MAML}}\xspace}
\newcommand{\pmaml}{{{P-MAML}}\xspace}

\newcommand{\fomaml}{{{FOMAML}}\xspace}
\newcommand{\imaml}{{{iMAML}}\xspace}
\newcommand{\maml}{{{MAML}}\xspace}
\newcommand{\rap}{{{Reptile}}\xspace}
\newcommand{\cmt}[1]{}
\newcommand{\eg}{{\textit{e.g.},}\xspace}
\newcommand{\ie}{{\textit{i.e.},}\xspace}
\newcommand{\etc}{{\textit{etc}.}\xspace}

\newcommand{\zsdc}[1]{{#1}}

\newcommand{\akil}[1]{{\leavevmode\color{red}{#1}}}
\renewcommand{\akil}[1]{}

\firstpageno{1}

\begin{document}

\title{Meta-Learning with Adjoint Methods}

\author{\name Shibo Li \email shibo@cs.utah.edu \\
       \addr School of Computing\\
       University of Utah
       \AND
       \name Zheng Wang \email wzhut@cs.utah.edu\\
       \addr School of Computing\\
       University of Utah
       \AND
       \name Akil Narayan \email akil@sci.utah.edu\\
       \addr Department of Mathematics, Scientific Computing and Imaging Institute\\
       University of Utah
       \AND
       \name Robert M. Kirby \email kirby@cs.utah.edu \\
       \addr School of Computing, Scientific Computing and Imaging Institute\\
       University of Utah
       \AND
       \name Shandian Zhe \email zhe@cs.utah.edu \\
       \addr School of Computing\\
       University of Utah }


\maketitle

\begin{abstract}
	 Model Agnostic Meta Learning (MAML) is widely used to find a good initialization for a family of tasks. Despite its success, a critical challenge in MAML is to calculate the gradient w.r.t. the initialization of a long training trajectory for the sampled tasks, because the computation graph can rapidly explode and the computational cost is very expensive. 
	 To address this problem, we propose Adjoint MAML (A-MAML). We view gradient descent in the inner optimization as the evolution of an Ordinary Differential Equation (ODE). To efficiently compute the gradient of the validation loss w.r.t. the initialization, we use the adjoint method to construct a companion, backward ODE. To obtain the gradient w.r.t. the initialization, we only need to run the standard ODE solver twice --- one is forward in time that evolves a long trajectory of gradient flow for the sampled task; the other is backward and solves the adjoint ODE. We need not create or expand any intermediate computational graphs, adopt aggressive approximations, or impose proximal regularizers in the training loss.  Our approach is cheap, accurate, and adaptable to different trajectory lengths. We demonstrate  the advantage of our approach in both synthetic and real-world meta-learning tasks. 
\end{abstract}

\section{Introduction}
Meta-learning paradigms~\citep{schmidhuber1987evolutionary, thrun2012learning} intend to  develop methods that can quickly adapt a learning model to new tasks or environments, like human learning. A prominent example is the recent model-agnostic meta-learning (\maml) algorithm~\citep{finn2017model}, which is particularly successful in learning the model initialization for a family of tasks. \maml is a bi-level optimization approach. The inner level starts from the initialization, and optimizes the training loss of the sampled tasks via gradient descent. At the trained model parameters, the outer-level uses back-propagation to calculate the gradient of the validation loss w.r.t. the initialization, and optimizes the initialization accordingly. 

While successful, a critical challenge of \maml is to back-propagate the gradient from a long training trajectory of the sampled tasks, because the resulting computation graph grows quickly, can easily explode, and is computationally expensive.\cmt{the resulting optimization also suffers from the vanishing gradient problem.} To combat these issues, practical usage of \maml performs only one or a few steps of gradient descent in the inner optimization; unfortunately this propagates a trajectory only close to the initialization, and fails to reflect the longer-term learning performance of using that initialization. To bypass this issue, first-order \maml (\fomaml)~\citep{finn2017model} and \rap ~\citep{nichol2018first} employ dropout on the Jacobian to obtain an aggressive approximation. While this is efficient, the approach loses accurate gradient information. The recent \imaml approach~\citep{rajeswaran2019meta} uses an implicit method to calculate an accurate gradient w.r.t. the initialization. This approach is elegant and successful, but imposes several restrictions. First, an additional regularizer that encourages proximity of the model parameters and the initialization must be added into the training loss. Second, the gradient is accurate only when training reaches the optimum of the regularized loss. 
\cmt{
\setlength{\columnsep}{5pt}
\begin{wrapfigure}{r}{0.5\textwidth}
	\centering
	\includegraphics[width=0.5\textwidth]{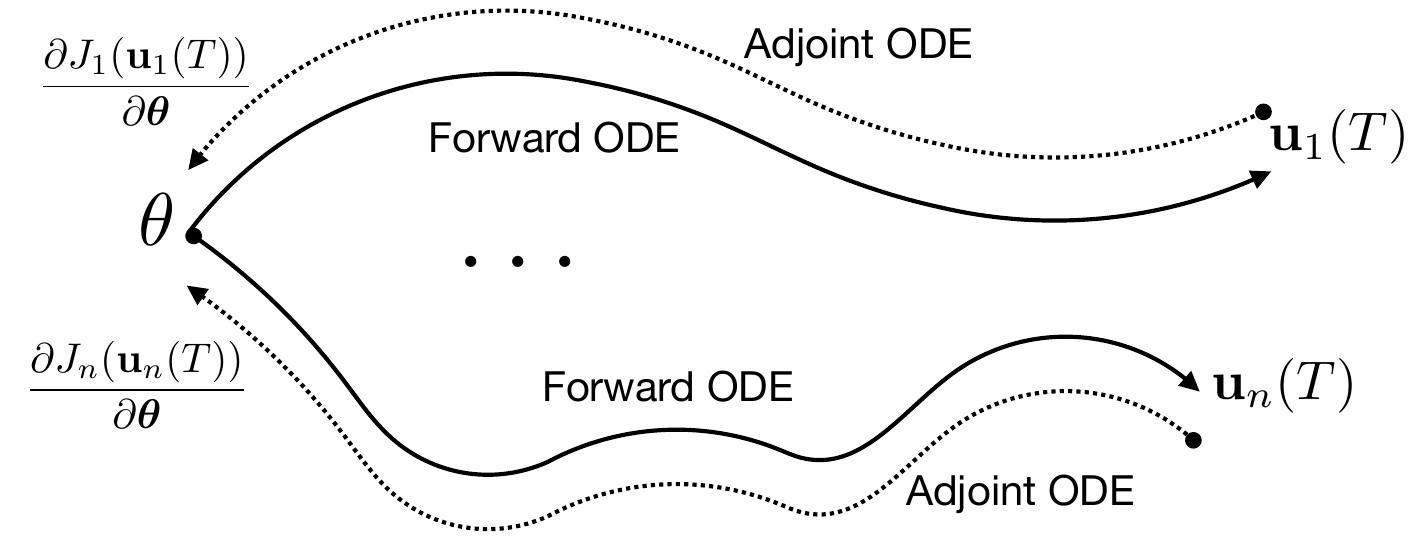}
	\caption{\small  Illustration of \ours, where $\btheta$ is the initialization, $J_n$ is the validation loss for task $n$ ($n=1, 2, \ldots$),  $\u_n$ are the model parameters for task $n$, and also the state of the corresponding forward ODE. \ours solves the forward ODE to optimize the meta-training loss, and solves the adjoint ODE backward to obtain the gradient of the meta-validation loss w.r.t. $\btheta$. } 	
	\label{fig:amaml}
	\vspace{-0.15in}
\end{wrapfigure}
}
\begin{figure}
	\centering
	\includegraphics[width=0.7\textwidth]{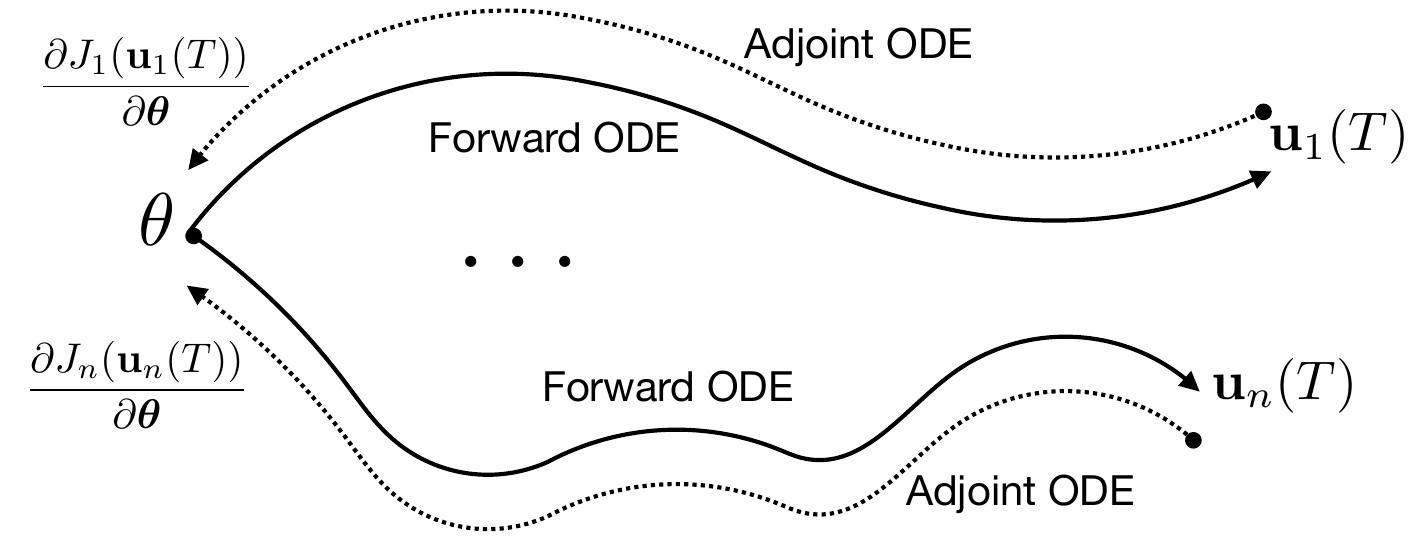}
	\caption{\small  Illustration of \ours, where $\btheta$ is the initialization, $J_n$ is the validation loss for task $n$ ($n=1, 2, \ldots$),  $\u_n$ are the model parameters for task $n$, and also the state of the corresponding forward ODE. \ours solves the forward ODE to optimize the meta-training loss, and then solves the adjoint ODE backward to obtain the gradient of the meta-validation loss w.r.t. $\btheta$. } 	
	\label{fig:amaml}
\end{figure}

In this paper, we propose \ours, an efficient and accurate approach to differentiate long paths of the inner-optimization in meta-learning. See Fig. \ref{fig:amaml} for an illustration. Our method does not require additional regularizers and can adapt to different trajectory lengths, hence it is well suited to commonly used training strategies, such as early stopping.  Specifically, we view the inner optimization (training) as evolving a forward ordinary differential equation (ODE) system, where the states are the model parameters. The standard gradient descent is equivalent to solving this ODE with the forward Euler method.  To calculate the gradient of the validation loss w.r.t. the model initialization, \ie the initial state of the ODE, we use the adjoint method to construct a companion  ODE. In effect, we only need to run the standard ODE solver twice: First, we solve the forward ODE to evolve a long training trajectory, based on which we compute the initial state of the adjoint ODE. Next, we solve the adjoint ODE backward to obtain the gradient w.r.t. the model initialization. To avoid divergence when solving backward, we use high-order solvers in the forward pass and track the states in the trajectory, based on which we use the modified Euler method (second-order) to solve backward. Throughout the procedure, we do not create and grow any intermediate computation graphs, nor do we apply any gradient approximation. The memory cost is linear in the number of model parameters. The accuracy is determined by the numerical precision of the ODE solver, which we can explicitly trade for speed.

For evaluation, we first examined \ours in two synthetic benchmark tests, regressing Alpine and Cosine mixture functions. In both task populations, we examined, starting from the given initialization, how the prediction error of the target model varies along with the increase of training epochs. \ours leads to much better prediction accuracy and training behavior compared against \maml, \fomaml, \rap, and \imaml.  Meanwhile, \ours dramatically reduces the memory usage and can easily scale to long training trajectories, compared with \maml which utilizes computation graphs.   The running time of \ours is comparable to \fomaml, \rap, and \imaml. We then applied \ours in three real-world applications of collaborative filtering and two image-classification tasks.  In several few-shot learning settings, \ours nearly always provides the best initialization, which leads to smaller prediction errors than the competing approaches during the meta-tests. The improvement is often significant. 

\section{Preliminaries}\label{sect:bg}
Suppose we have a family of correlated learning tasks $\Acal $. The size of $\Acal$ can be very large or even infinite. For each task, we use the same machine learning model $\Mcal$, which is parameterized by $\u \in \mathds{R}^d$, \eg a deep neural network. Our goal is to learn an initialization $\btheta$  for $\u$, which can well adapt to all the tasks in $\Acal$. To this end, we sample $N$ tasks, $\Scal = \{\Tcal_{1}, \ldots, \Tcal_{N}\}$, from a task distribution $p$ on $\Acal$, and for each $\Tcal_{n}$, we collect a dataset $\Dcal_n$. We use the $N$ datasets $\widehat{\Dcal} = \{\Dcal_1, \ldots, \Dcal_N\}$ to meta-learn $\btheta$. We expect that given any new task $\Tcal^* \in \Acal$, after initializing  $\u$ with $\btheta$, the training of $\Mcal$ on $\Tcal^{*}$ can achieve better performance with the same or fewer training epochs or iterations or examples.

A particularly successful meta-learning algorithm is model-agnostic meta-learning (\maml)~\citep{finn2017model}, which uses a bi-level optimization approach to estimate $\btheta$. Specifically, each  $\Dcal_n$ is partitioned into a meta-training dataset $\Dcal_n^{\text{tr}}$ and a meta-validation dataset $\Dcal_n^{\text{val}}$. In the inner level, we start with $\btheta$ and optimize the training loss $\Lcal(\u, \Dcal_n^{\text{tr}})$ for each task $n$. Let us denote the trained parameters by $\bpsi_n(\btheta)$. In the outer level, we evaluate these trained parameters on the validation loss, and optimize $\btheta$ accordingly, \ie $\btheta^* = \min\; \frac{1}{N} \sum\nolimits_{j=1}^N \Lcal(\bpsi_n(\btheta), \Dcal_n^{\text{val}})$. 
\maml obtains the gradient w.r.t. $\btheta$ via automatic differentiation, which essentially computes $\frac{\d \bpsi_n(\btheta)}{\d \btheta}$ via back-propagation on a computation graph. However, this is very challenging for long training trajectories to obtain $\bpsi_n(\btheta)$, since the computation graph can rapidly explode and become very expensive to compute.\cmt{, and also suffer from the vanishing gradient problem. } Therefore, in practice, \maml typically only conducts one or a few gradient descent steps in the inner optimization, \eg with one step, $\bpsi_n(\btheta) = \btheta - \alpha \nabla \Lcal(\btheta, \Dcal_n^{\text{tr}})$, 
where $\alpha$ is the step size.\akil{Use ``learning rate" instead of ``step size"? I assume the former is more common in the ML community?} However, with only one step the obtained parameters are frequently too close to the initialization, and inadequately reflect the actual longer-range training performance.

To bypass this issue, First-Order \maml (\fomaml)~\citep{finn2017model} drops out the Jacobian $\frac{\d \bpsi_n(\btheta)}{\d \btheta}$ and replaces it with the identity matrix $\I$. In so doing, \fomaml can perform many gradient descent steps to obtain $\bpsi_n$ and  update $\btheta$ with
\[
\btheta \leftarrow \btheta - \eta \cdot \frac{1}{N}\sum\nolimits_{n=1}^N \frac{\partial \Lcal(\bpsi_n, \Dcal_n^{\text{val}})}{\partial \bpsi_n},
\]
where $\eta$ is the learning rate. With the same idea, \rap~\citep{nichol2018first} instead adjusts the updating direction to $\frac{1}{N}\sum_{j=1}^N \frac{\partial \Lcal(\bpsi_n, \Dcal_n^{\text{val}})}{\partial \bpsi_n} - \btheta$. Despite being efficient, these methods lack accurate gradient information about $\btheta$. To overcome this limitation, the recent work, \imaml~\citep{rajeswaran2019meta}, calculates the accurate gradient via an implicit gradient method. However, it needs to incorporate a proximity regularizer into the training loss to bind $\u$ and $\btheta$ explicitly,  
\[
\widehat{\Lcal}(\u, \Dcal_n^{\text{tr}}) = \Lcal(\u, \Dcal_n^{\text{tr}}) + \frac{\lambda}{2} \| \u - \btheta\|^2.
\]
The accurate gradient can be obtained (only) when the training reaches the optimum, \ie $\bpsi_n = \argmin_{\u} \widehat{\Lcal}(\u, \Dcal_n^{\text{tr}}) $, since  we can derive the implicit gradient $\frac{\d \bpsi_n}{\d \btheta}$ from the fact  that $\frac{\partial \widehat{\Lcal}}{\partial \bpsi_n} = \0$.

\section{Adjoint MAML}
In this paper, we propose \ours, which can accurately and efficiently compute the gradient of the meta loss w.r.t. the initialization for long training trajectories, without the need for aggressive approximations or additional regularization, and adapts to different trajectory lengths. Hence, our method can be easily integrated with common training strategies, \eg early stopping. 
\subsection{ODE View of Inner Optimization}
Specifically, we first view the inner optimization as evolving an ODE system. In more detail, given task $n$,  starting from $\btheta$, we run gradient descent for a long time to train the model. The training procedure can be in more general viewed as solving the following ODE,
\begin{equation}
	\begin{cases}
	\u_n(0) &= \btheta, \notag \\
	\frac{\d \u_n}{\d t} &= -\frac{\partial \Lcal(\u_n, \Dcal_n^{\text{tr}})}{\partial \u_n},
	\end{cases} 
\end{equation}
where the state $\u_n(t)$ represents the model parameters at time $t$. Running gradient descent with a step size 
\akil{again, ``learning rate"?}
$\alpha$ essentially solves the ODE with the forward Euler method using temporal step size $\alpha$, 
corresponding to the update $\u_n(t+\alpha) \leftarrow \u_n(t) - \alpha \frac{\partial \Lcal(\u_n, \Dcal_n^{tr})}{\partial \u_n}$. However, the ODE view allows us to apply a variety of more efficient, high-order solvers to fulfill the training, \eg the Runge-Kutta method~\citep{dormand1980family}. Suppose we stop at time $T$, then we evaluate the trained parameters $\u_n(T)$ on the validation dataset via $\Lcal(\u_n(T), \Dcal_n^{val})$. Therefore, the meta loss is given by
\begin{align}
	J(\btheta) = \frac{1}{N}\sum\nolimits_{n=1}^N \Lcal(\u_n(T), \Dcal_n^{\text{val}}). \label{eq:meta-loss-ode}
\end{align}
Note that the stopping time $T$ is not necessarily the same for all the tasks; it can vary for different tasks as determined, say, by an early stopping criterion. 

\subsection{Efficient Back-Propagation via Solving Adjoint ODEs}
To optimize $\btheta$ in \eqref{eq:meta-loss-ode} (in the outer loop), we need to be able to compute the gradient of the  validation loss for each task $n$, \ie $\frac{\d J_n}{\d \btheta}$, where $J_n = \Lcal(\u_n(T), \Dcal_n^{\text{val}})$. We seek to compute this gradient efficiently for large $T$ without creating and growing a computation graph.
To this end, we use the adjoint method~\citep{pontryagin1987mathematical}. 
To simplify the notation, we first define
\begin{align}
	J_n(\u_n(T)) &= \Lcal(\u_n(T), \D_n^{\text{val}}), \notag \\
	f(\u_n, \Dcal_n^{tr}) &= -\left(\frac{\partial \Lcal(\u_n, \Dcal_n^{\text{tr}})}{\partial \u_n}\right)^\top. \label{eq:def}
\end{align}
Note that we use the row vector representation of the gradient, \ie $\frac{\partial \Lcal}{\partial \u_n}$ is a $1 \times d$ vector. This is consistent with the shape of Jacobian matrix, and the chain rule can be expressed as the matrix multiplication from left to right, which is natural and convenient. Accordingly, the ODE for $\u_n(t)$ can be written as 
\begin{equation}
	\begin{cases}
			\u_n(0) &= \btheta,  \\
		\frac{\d \u_n}{\d t} &= \f(\u_n, \D_n^{tr}). \label{eq:forward-ode}
	\end{cases}
\end{equation}

Next, to construct an adjoint ODE for efficient gradient computation, we augment the validation loss,
\begin{align}
	\whJ_n = J_n\left(\u_n(T)\right) + \int_0^T \blambda(t)^\top \left(f(\u_n, \D_n^{\text{tr}}) - \frac{\d \u_n}{\d t}\right) \d t,  \label{eq:aug-J}
\end{align}
where $\blambda(t)$ is a Lagrange multiplier and a $d \times 1$ vector. According to the ODE constraint \eqref{eq:forward-ode},  the extra integral in \eqref{eq:aug-J} is $0$ and $\whJ_n = J_n$. Hence, we have 
\begin{align}
	&\frac{\d J_n}{\d \btheta} = \frac{\d \whJ_n}{\d \btheta} = \frac{\partial J_n}{\partial \u_n(T)} \frac{\d \u_n}{\d \btheta}(T) \notag \\
	&+ \int_0^T \blambda^\top \left[\frac{\partial \f}{\partial \u_n} \frac{\d \u_n}{\d \btheta} - \frac{\d \frac{\d \u_n}{\d t}}{\d \btheta}\right] \d t.  \label{eq:grad}
\end{align}
For the second term in the integral, we switch the derivative order and apply integration by parts, 
	\begin{align}
		&\int_0^T \blambda^\top  \frac{\d \frac{\d \u_n}{\d t}}{\d \btheta} \d t = \int_0^T \blambda^\top  \frac{\d \frac{\d \u_n}{\d \btheta}}{\d t} \d t \notag \\
		&= \blambda^\top \frac{\d \u_n}{\d \btheta}\bigg|_0^T - \int_0^T \left(\frac{\d \blambda}{\d t}\right)^\top \frac{\d \u_n}{\d \btheta} \d t \notag \\
		&= \blambda(T)^\top \frac{\d \u_n}{\d \btheta}(T) - \blambda(0)^\top \frac{\d \u_n}{\d \btheta}(0) - \int_0^T \left(\frac{\d \blambda}{\d t}\right)^\top \frac{\d \u_n}{\d \btheta} \d t. \notag 
	\end{align}
Substituting the above into \eqref{eq:grad}, we obtain  
\begin{align}
	\frac{\d J_n}{\d \btheta}  &= \frac{\partial J_n}{\partial \u_n(T)} \textcolor{blue}{\frac{\d \u_n}{\d \btheta}(T)} - \blambda(T)^\top \textcolor{blue}{\frac{\d \u_n}{\d \btheta}(T)} + \blambda(0)^\top \frac{\d \u_n}{\d \btheta}(0) \notag \\
	&+ \int_0^T\left\{ \blambda^\top \frac{\partial \f}{\partial \u_n}\textcolor{blue}{\frac{\d \u_n}{\d \btheta}} + \left(\frac{\d \blambda}{\d t}\right)^\top \textcolor{blue}{\frac{\d \u_n}{\d \btheta}}\right\}\d t. \notag 
\end{align}
The computationally expensive term is the Jacobian $\frac{\d \u_n}{\d \btheta}$ (marked as blue), which we efficiently handle by constructing an adjoint ODE for the Lagrange multiplier $\blambda$,   
\begin{align}
	\begin{cases}
	\blambda(T) &= \left(\frac{\partial J_n}{\partial \u_n(T)}\right)^\top, \\
	 	\left(\frac{\d \blambda}{\d t}\right)^\top& = - \blambda(t)^\top \frac{\partial \f}{\partial \u_n}. \label{eq:adjoint-ode}
\end{cases}
\end{align}

Note that the ODE \eqref{eq:adjoint-ode} runs backward in time starting at the terminal time $T$. If we can solve \eqref{eq:adjoint-ode}, the Jacobian terms (blue) will cancel, and the full gradient becomes 
\begin{align}
	\frac{\d J_n}{\d \btheta} = \blambda(0)^\top \frac{\d \u_n}{\d \btheta}(0) =  \blambda(0)^\top, 
\end{align}
where we have used $\frac{\d \u_n}{\d \btheta}(0) = \I$. We see that the gradient is simply the state of $\blambda$ at time $0$. To confirm the feasibility of solving \eqref{eq:adjoint-ode}, we can see from  \eqref{eq:adjoint-ode} and \eqref{eq:def} that $\frac{\partial \f}{\partial \u_n} = \H({\u_n}) = -\frac{\partial^2 \Lcal(\u_n, \Dcal_n^{tr})}{\partial \u_n^2}$
 is the Hessian matrix of the model parameters. While it seems extremely costly to calculate the Hessian, when we substitute the above Hessian into \eqref{eq:adjoint-ode} and  take the transpose, we find,
 \begin{align}
 	\begin{cases}
 	\blambda(T) &= \left(\frac{\partial J_n}{\partial \u_n(T)}\right)^\top, \\
 	 \frac{\d \blambda}{\d t} &= - \H(\u_n) \blambda(t). \label{eq:adjoint-ode-v2}
 	\end{cases}
 \end{align}
Now it is clear that the dynamics of $\blambda$ is a Hessian-vector product. It is known that we never need to explicitly compute the Hessian matrix. We can first compute the gradient $\g =\frac{\partial \Lcal}{\partial \u_n}$,  then the dot product $s = \v^\top \g$, and take the gradient of the scalar $s$ again, which gives exactly $\H\blambda$.  The complexity is the same as computing the gradient.

Therefore, to calculate $\frac{\d J_n}{\d \btheta}$, we only need to run standard ODE solvers twice. First, we run a solver to evolve \eqref{eq:forward-ode} from time $0$ to time $T$. Note that even a small $T$ can correspond to many gradient descent steps. For example, $T = 10$ corresponds to running $1000$ gradient descent steps where the step size is set to $0.01$ (a common choice). We can apply high-order methods, like RK45~\citep{dormand1980family} to further improve the speed and accuracy. Next, at the trained parameters $\u(T)$, we jointly solve \eqref{eq:adjoint-ode-v2} and \eqref{eq:forward-ode} backward (note that dynamics of $\blambda$ needs $\u_n$). For solving both ODEs, we never need to create and/or grow  new computation graphs. All we need is to compute the dynamics in \eqref{eq:forward-ode} and \eqref{eq:adjoint-ode-v2}, and the computational complexity is the same as computing the gradient of the training loss w.r.t the model parameters. The memory cost only involves storage of $\u_n$ and $\blambda$, which is proportional to the number of model parameters. We never need to maintain or calculate any Jacobian matrix. The accuracy is determined by the numerical precision of the ODE solvers, which have been developed for decades, are mature, and can easily effect tradeoffs between precision and speed.  Note that our method does not need to add extra regularization into the training loss, although our framework can be easily adjusted to support such regularization. 

Empirically, we found that back-solving can diverge when $T$ is very large, say, $100$. {This might be because a larger $T$ increases the chance that different forward trajectories (\ie starting from different initial states) intersect or even overlap. This is not uncommon for gradient-based training --- even with different initializations, it might still arrive at or explore the same area. If so, when we solve the ODEs backward, it is easy to diverge at the intersection points.} To promote robustness, we track the state $\u_n$ in the training trajectory with a given step size during the forward solve. This can be automatically done via the ODE solver. Then based on the list of states $\{\u_{n,j}\}_j$, we solve the adjoint ODE backward with the modified Euler method~\citep{ascher1998computer} whose global accuracy is $\Ocal(h^2)$ where $h$ is the ODE solver step size.  Specifically, at each step $j$, we first calculate an intermediate value $\widetilde{\blambda}_j$ and then the state $\blambda_j$ via,
\begin{align}
	\widetilde{\blambda}_{j} &= \blambda_{j+1} + h \H(\u_{n,j+1})\blambda_{j+1}, \notag \\
	 \blambda_j &= \blambda_{j+1} + \frac{h}{2}\left[\H(\u_{n,j+1})\blambda_{j+1} + \H(\u_{n,j})\widetilde{\blambda}_{j}\right]. \notag 
\end{align}
While this increases memory requirements, it is still linear with the number of parameters, $\Ocal(\frac{T}{h} d)$, and much cheaper than building a computational graph. The experiments show that our method can  scale to long training trajectories very economically (see Sec. \ref{sect:memory}). Our method is summarized in Algorithm \ref{alg:amaml}. 

\begin{algorithm}                  
	\small
	\caption{\ours ($p(\Tcal)$, $T$, $\eta$, $G$, $\xi$)}          
	\label{alg:amaml}                           
	\begin{algorithmic}[1]                    
		\STATE Randomly initialize $\btheta$.
		\REPEAT 
		\STATE Sample a mini-batch of tasks $\{\Tcal_n\}_{n=1}^B$ from $p(\Tcal)$.
		\FOR {each task $\Tcal_n$}
		\STATE Calculate $\frac{\partial J_n}{\partial \btheta}$ with Algorithm \ref{alg:adjoint}.
		\ENDFOR
		\STATE $\btheta \leftarrow \btheta - \eta \cdot \frac{1}{B} \sum_{n=1}^B \frac{\partial J_n}{\partial \btheta}$ (or use ADAM).
		\UNTIL{$G$ iterations are done or the change of $\btheta$ is less than $\xi$}
		\STATE Return $\btheta$. 
	\end{algorithmic}
\end{algorithm}
\begin{algorithm}
	\small 
	\caption{Adjoint Gradient Computation ($\btheta$, $J_n$, $T$, $h$)}          
	\label{alg:adjoint}                           
	\begin{algorithmic} [1]                  
		\STATE $\u_n(0) \leftarrow \btheta$.
		\STATE Solve forward ODE \eqref{eq:forward-ode} to time $T$ with RK45, and track the states $\{\u_{n,j}\}_j$ in the trajectory with step size $h$.
		\STATE $\blambda(T) \leftarrow \frac{\partial J_n}{\partial \u_n(T)}$.
		\STATE Solve the adjoint ODE \eqref{eq:adjoint-ode-v2} to time $0$ with modified Euler method based on the state list $\{\u_{n,j}\}$.
		\STATE Return  $\blambda(0)$.
	\end{algorithmic}
\end{algorithm}

\section{Related Work}
Meta-learning \citep{schmidhuber1987evolutionary, thrun2012learning, naik1992meta} \cmt{[51,55, 41]} can be (roughly) classified into three categories: (1) metric-learning methods that learn a metric space (in the outer lever), where the tasks (in the inner level) make predictions by simply matching the training points, \eg nonparametric nearest neighbors \citep{koch2015siamese, vinyals2016matching, snell2017prototypical, oreshkin2018tadam, allen2019infinite}\cmt{[29, 57, 54, 45, 3]}, (2) black-box methods that train feed-forward or recurrent NNs to take the hyperparameters and task dataset as the input and  outright predict the optimal model parameters or parameter updating rules \citep{hochreiter2001learning, andrychowicz2016learning, li2016learning, ravi2016optimization, santoro2016meta, duan2016rl, wang2016learning, munkhdalai2017meta, mishra2017simple}\cmt{[25, 5, 33, 48,50, 12, 58, 40, 38]}, and   (3) optimization-based methods that conduct a bi-level optimization, where the inner level is to estimate the model parameters given the hyperparameters (in each task) and the outer level is to optimize the hyperparameters via a meta-loss~\citep{finn2017model, finn2018learning, bertinetto2018meta, lee2019meta,zintgraf2019fast, li2017meta, finn2018probabilistic, zhou2018deep, harrison2018meta}\cmt{[15, 13, 8, 60, 34, 17, 59, 23]}. Other approaches include \citep{rusu2018meta, triantafillou2019meta}, \etc A successful application of meta-learning is few-shot learning, for which important models include \citep{lake2011one, vinyals2016matching,snell2017prototypical,sung2018learning}, to name a few.  
An excellent survey about meta-learning for neural networks is given in~\citep{hospedales2020meta}.

\maml~\citep{finn2017model} is a popular optimization-based meta-learning method. In addition to FOMAML and Reptile, there are many variants, such as probabilistic versions~\citep{grant2018recasting,yoon2018bayesian,finn2018probabilistic}, and ones improving reinforcement learning~\citep{song2020maml,liu2019taming}. Recently, \citet{denevi2020advantage,wang2020structured,denevi2021conditional} proposed conditional meta learning to leverage side information (when available) to learn task-specific initializations. 
The recent work of \citep{im2019model,xu2021meta} also introduces an ODE view for \maml. However, they use the ODE theory and methods to analyze/improve the outer level optimization, where the inner level still performs one step gradient descent as in standard \maml.  They do not consider long training trajectories in the inner level. \citet{im2019model} pointed out the \maml update is a special case of (second-order) Runge-Kutta gradients, and suggested using more refined nodes, weights and even higher-order updates. \citet{xu2021meta} showed that if the outer-level optimization of \maml is considered as solving an ODE, it enjoys a linear convergence rate for strongly convex task losses. Based on their analysis, they proposed a bi-phase algorithm to further reduce the cost and improve efficiency.  Our work uses the ODE view for inner-level optimization.  The adjoint method is a classical and popular framework to estimate the parameters of ODE or dynamic control models~\citep{chen2018neural,eichmeir2021adjoint}. If we use Euler method to solve the adjoint ODE, it reduces to the reverse mode differentiation method~\citep{bengio2000gradient,baydin2014automatic}, yet leaving first-order global accuracy ($\Ocal(h)$). Another excellent related work is ~\citep{domke2012generic} that provides a general bi-level optimization framework. It can optimize  hyper-parameters that explicitly show up in the loss (\eg regularization strength, yet not including the parameter initialization) with the inner-optimization procedure taken into account.
\section{Experiments}
\subsection{2D regression}\label{sect:syn}
\begin{figure*}[t]
	\centering
	\setlength\tabcolsep{0.01pt}
	\captionsetup[subfigure]{aboveskip=0pt,belowskip=0pt}
	\begin{tabular}[c]{ccc}
		\begin{subfigure}[t]{0.32\textwidth}
			\centering
			\includegraphics[width=\textwidth]{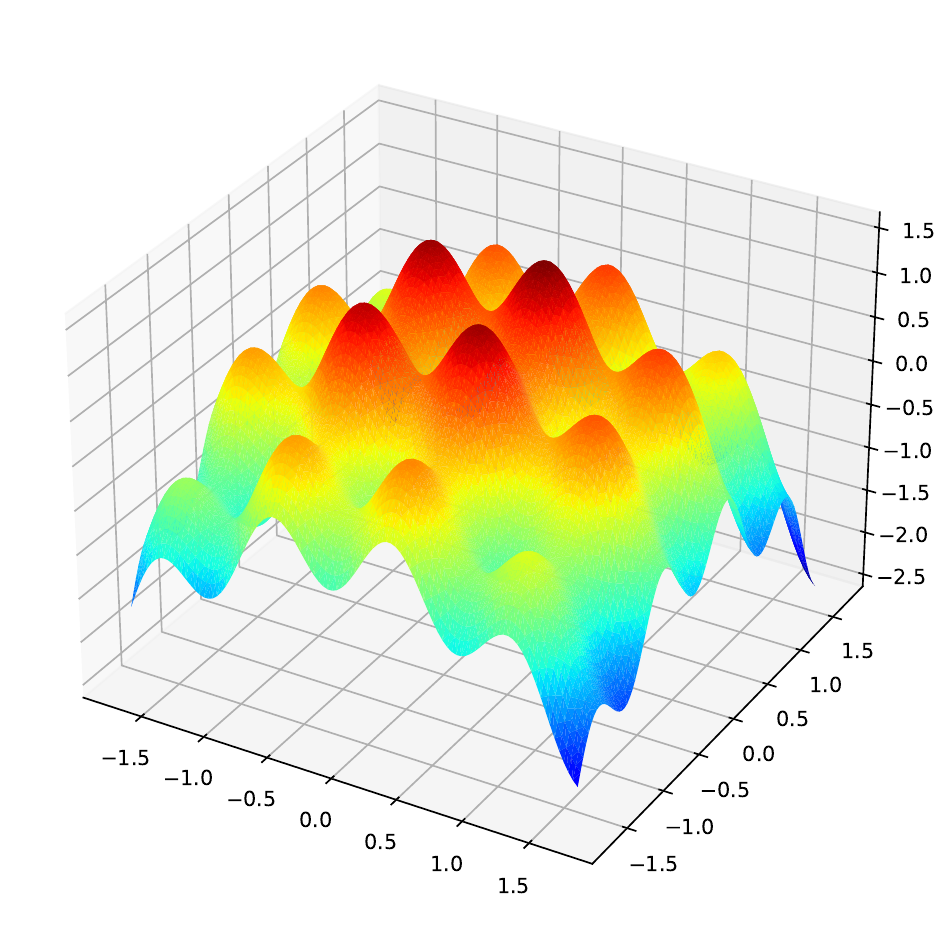}
			\caption{\small \textit{CosMixture} instance} \label{fig:CosMixture-ins}
		\end{subfigure} 
		&
		\begin{subfigure}[t]{0.32\textwidth}
			\centering
			\includegraphics[width=\textwidth]{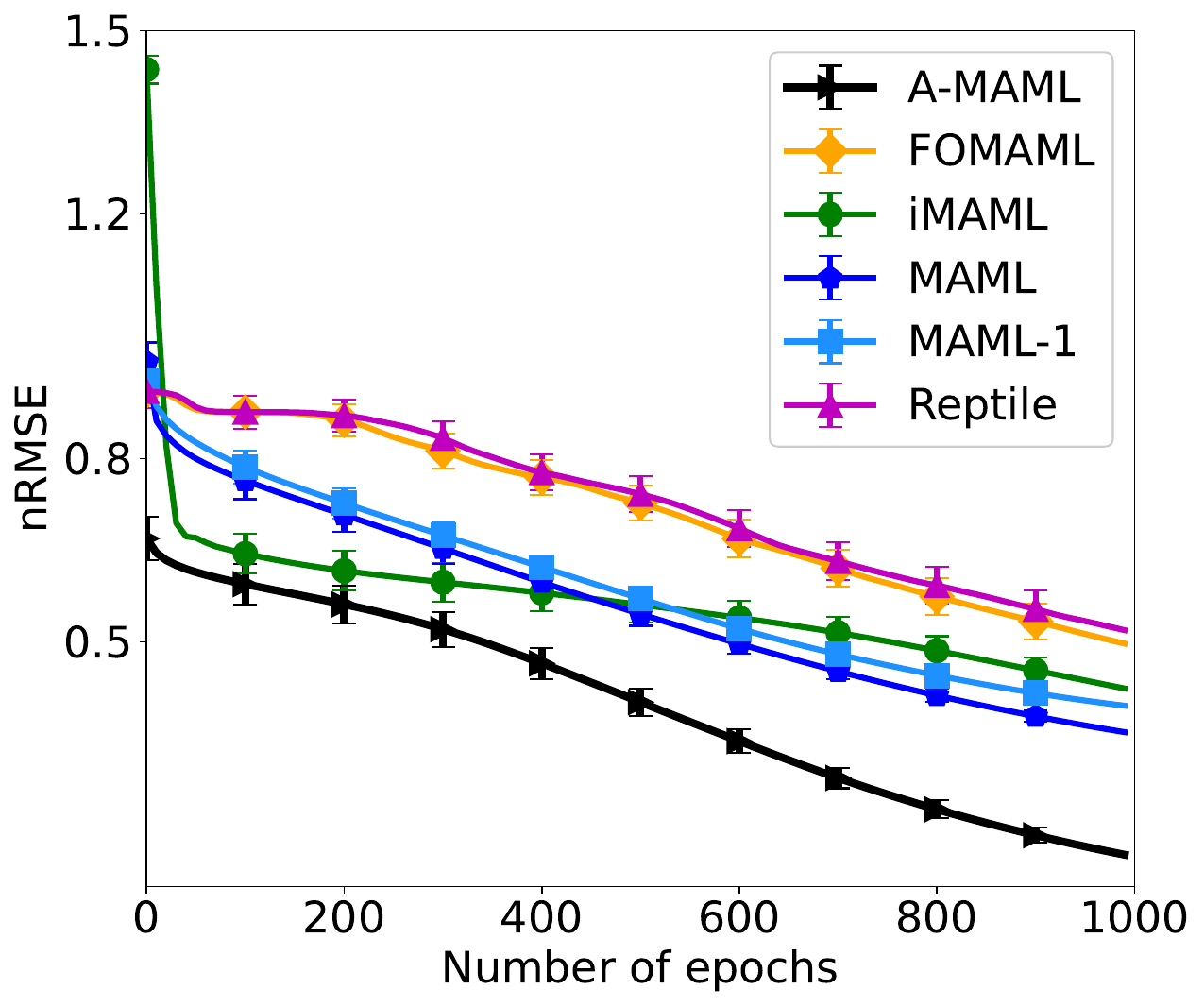}
			\caption{\small \textit{CosMixture}: 50shot-50val}
		\end{subfigure} 
		&
		\begin{subfigure}[t]{0.32\textwidth}
			\centering
			\includegraphics[width=\textwidth]{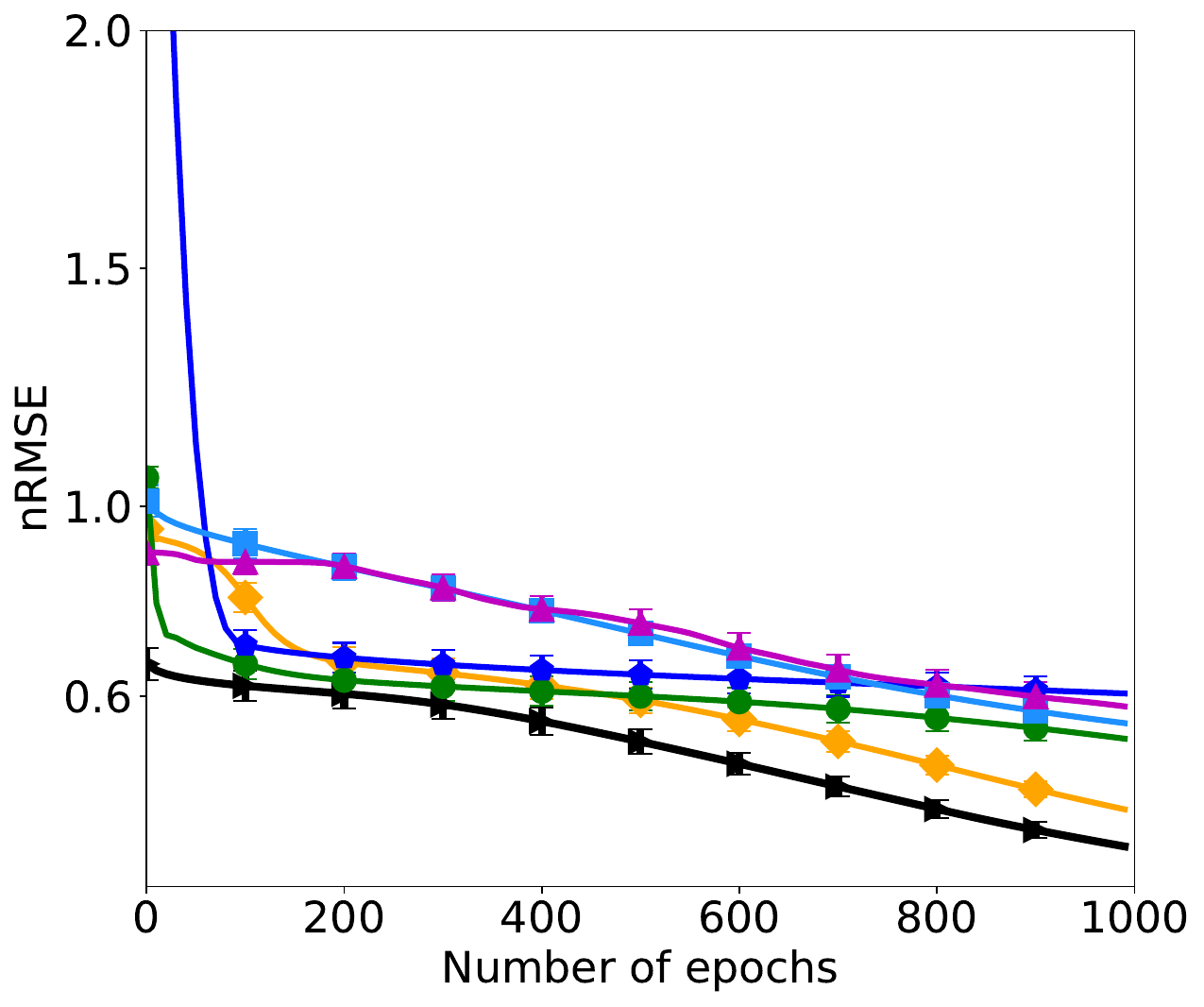}
			\caption{\small \textit{CosMixture}: 100shot-100val}
		\end{subfigure} 	\\
		\begin{subfigure}[t]{0.32\textwidth}
			\centering
			\includegraphics[width=\textwidth]{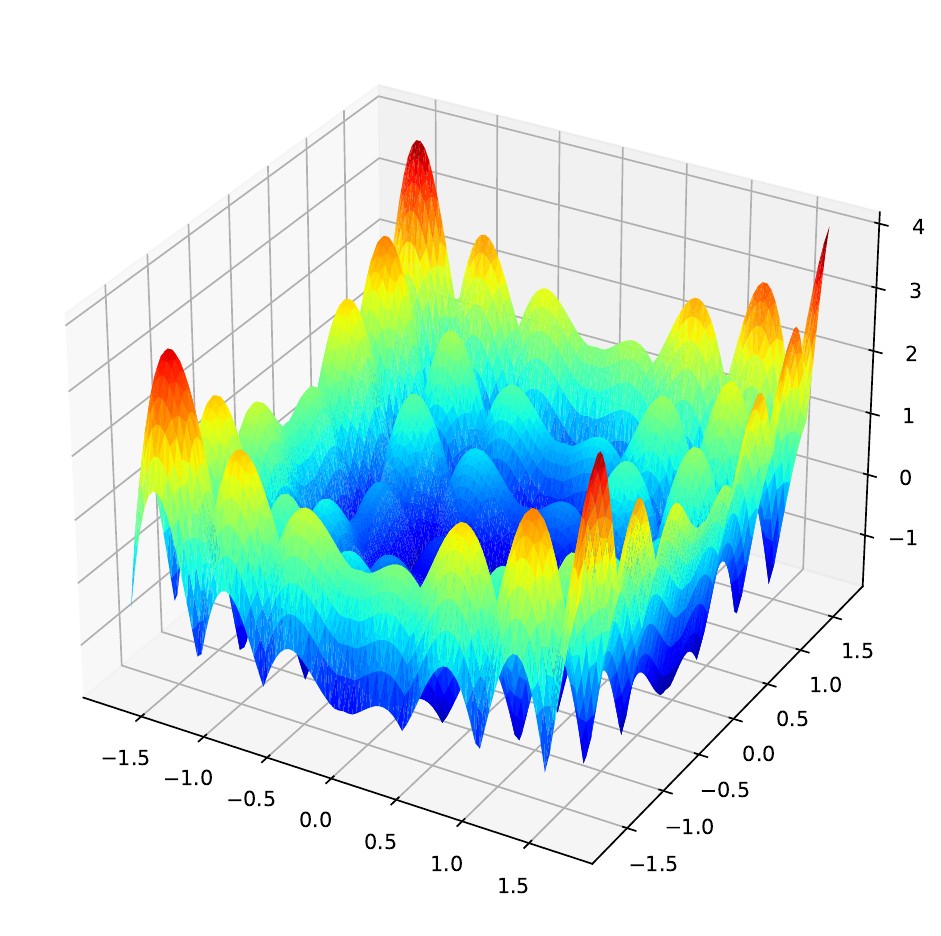}
			\caption{\small \textit{Alpine} instance} \label{fig:Alpine-ins}
		\end{subfigure} 
		&
		\begin{subfigure}[t]{0.32\textwidth}
			\centering
			\includegraphics[width=\textwidth]{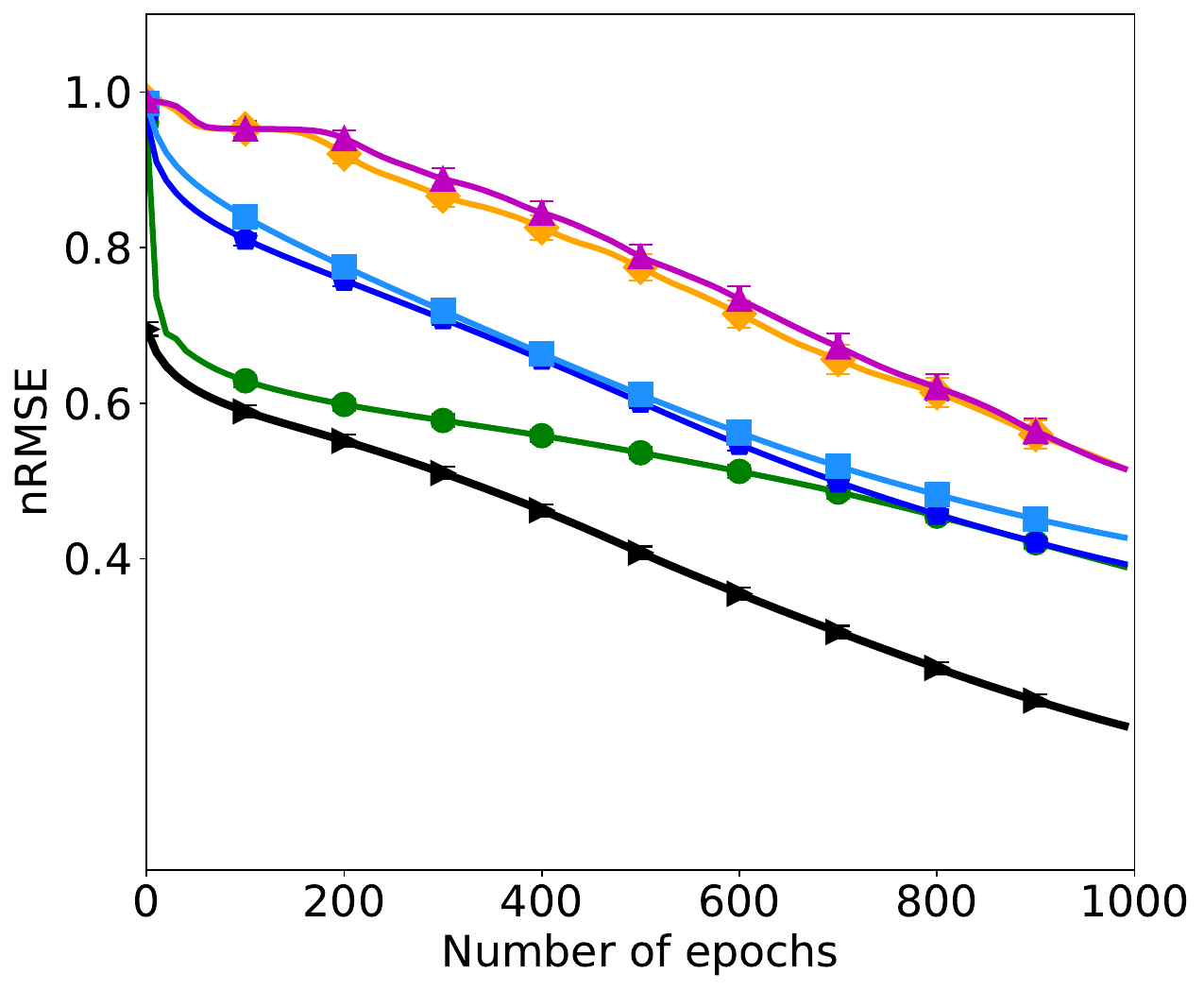}
			\caption{\small \textit{Alpine}: 50shot-50val}
		\end{subfigure} 
		&
		\begin{subfigure}[t]{0.32\textwidth}
			\centering
			\includegraphics[width=\textwidth]{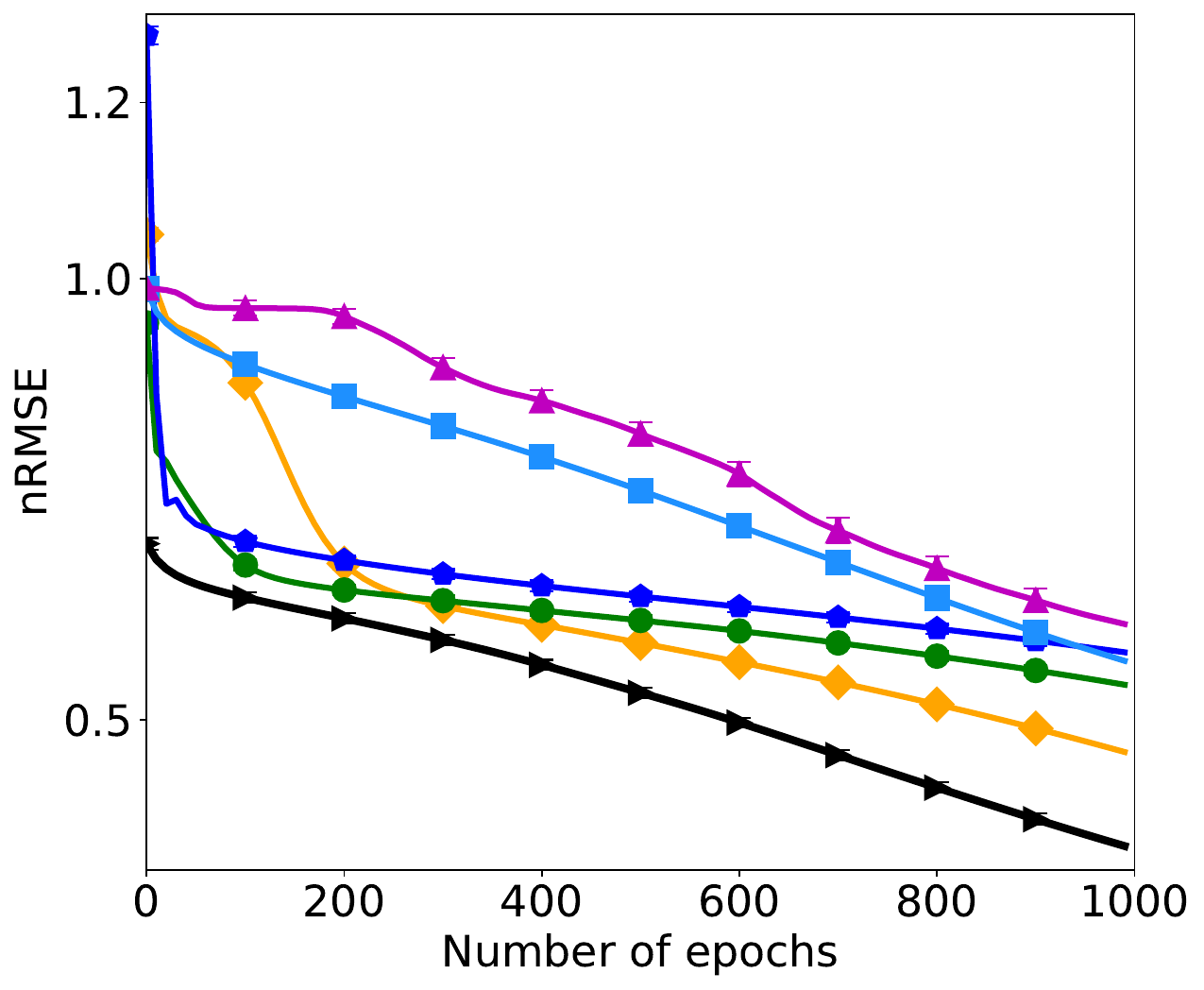}
			\caption{\small \textit{Alpine}: 100shot-100val}
		\end{subfigure} 
	\end{tabular}
	\caption{\small Prediction error of the neural network in learning \textit{CosMixture} and \textit{Alpine} function families, starting from the initialization provided by different meta-learning approaches. (a,d) are the instances of the two types of functions. 50shot-50val means 50 examples were used for meta-training and another 50 examples for meta-validation. 100shot-100val means both the meta-training and meta-validation used  100 examples. The results were averaged over 100 test tasks. } 	
	\label{fig:synthetic}
	\vspace{-0.1in}
\end{figure*}

\vspace{-0.05in}
For evaluation, we first examined the proposed approach in two synthetic benchmark tests, namely, meta learning of \textit{CosMixture} and \textit{Alpine} functions ({\url{http://infinity77.net/global_optimization/test_functions.html}}), both of which are 2D regression tasks. We considered two families of tasks. In the first family, each task aims to learn a specific \textit{CosMixture} function of the following form,
\begin{align}
	f_1(\x) = -0.1 \sum\nolimits_{i=1}^d A \cos(\omega x_i + \phi) - \sum\nolimits_{i=1}^d x_i^2,
\end{align}
where $\x \in [-1, 1]^2$, $d =2$, $A \in [0.1, 1.0]$, $\omega \in [0.5 \pi, 2.0\pi]$, and $\phi \in [3.0, 6.0]$. 
The second family of tasks learn instances of the \textit{Alpine} function,  
\begin{align}
	f_2(\x) = \sum\nolimits_{i=1}^d |x_i \sin(x_i + \phi_i) + 0.1x_i|,
\end{align}
where $\x \in [10, 10]^2$, $d=2$, $\phi_1 \in [-\frac{5}{12}\pi, \frac{5}{12}\pi]$, and $\phi_2 \in [-\frac{5}{12}\pi, \frac{5}{12}\pi]$.   An instance of each function is shown in Fig. \ref{fig:CosMixture-ins} and \ref{fig:Alpine-ins}. The learning model for both task populations is a neural network with two hidden layers, each consisting of 32 neurons with  Tanh activation. 
To conduct meta-learning for each task population, we randomly sampled $100$ tasks, where for each task, the parameters of the target function, \ie $\{A, \omega, \phi\}$ in \textit{CosMixture} and $\{\phi_1, \phi_2\}$ in \textit{Alpine}, are uniformly sampled from their ranges. We considered two meta-learning settings: 50shot-50val, where we used 50 examples for meta-training and 50 another examples in meta-validation, and 100shot-100val, where both the meta-training and meta-validation losses employed 100 examples.  These examples are non-overlapping and generated by uniformly sampling from the input domain. Given the learned initialization, we tested on $100$ new tasks, where the task training data were generated in the same way as in the meta-training and $100$ another examples were sampled to evaluate the prediction accuracy. 

\cmt{
\begin{figure*}[t]
	\centering
	\setlength\tabcolsep{0pt}
	\captionsetup[subfigure]{aboveskip=0pt,belowskip=0pt}
	\begin{tabular}[c]{cccc}
		\begin{subfigure}[t]{0.25\textwidth}
			\centering
			\includegraphics[width=\textwidth]{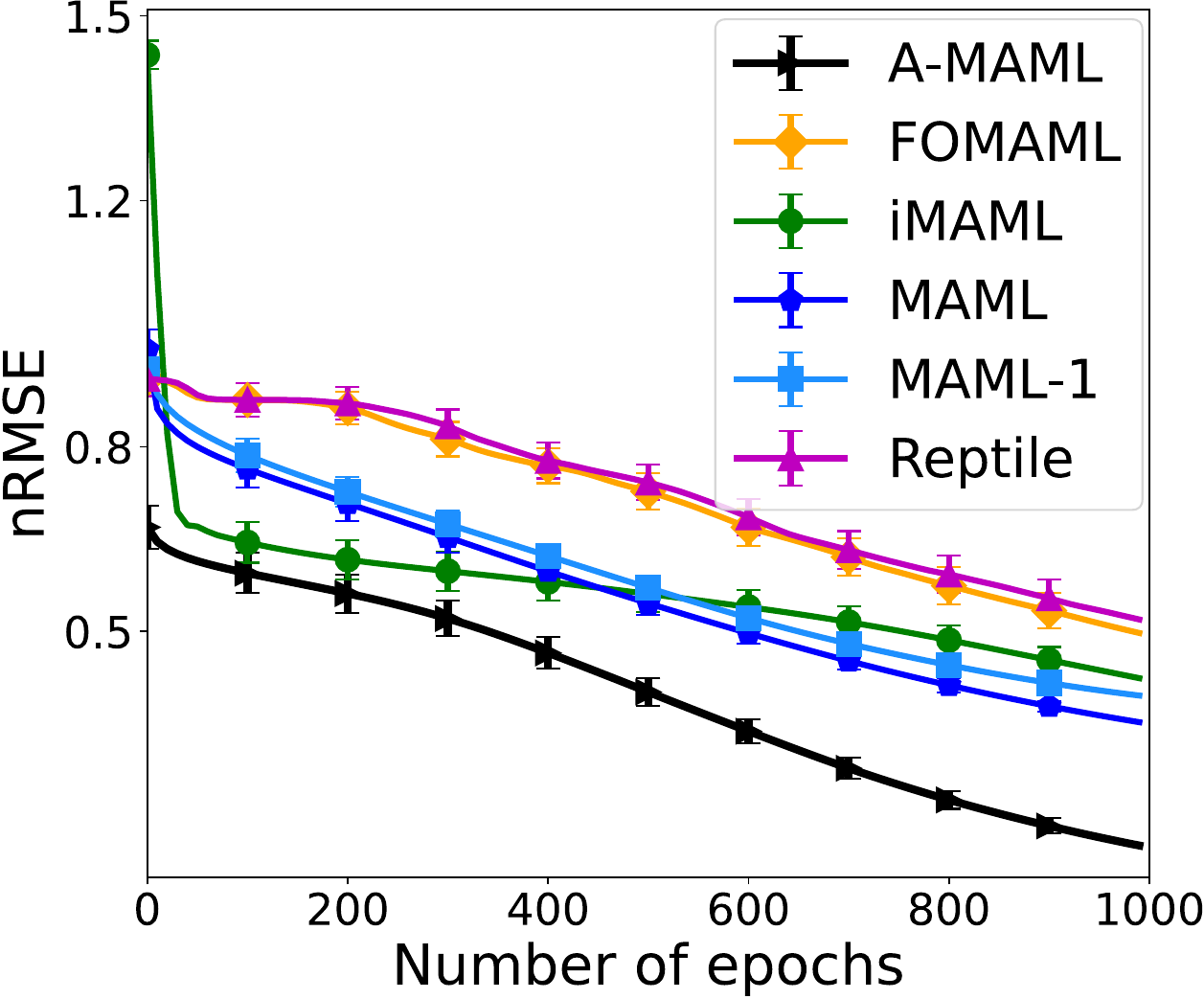}
			\caption{\small{ \textit{CosMixture}: 50shot-50val}}
		\end{subfigure} 
		&
	\begin{subfigure}[t]{0.25\textwidth}
		\centering
		\includegraphics[width=\textwidth]{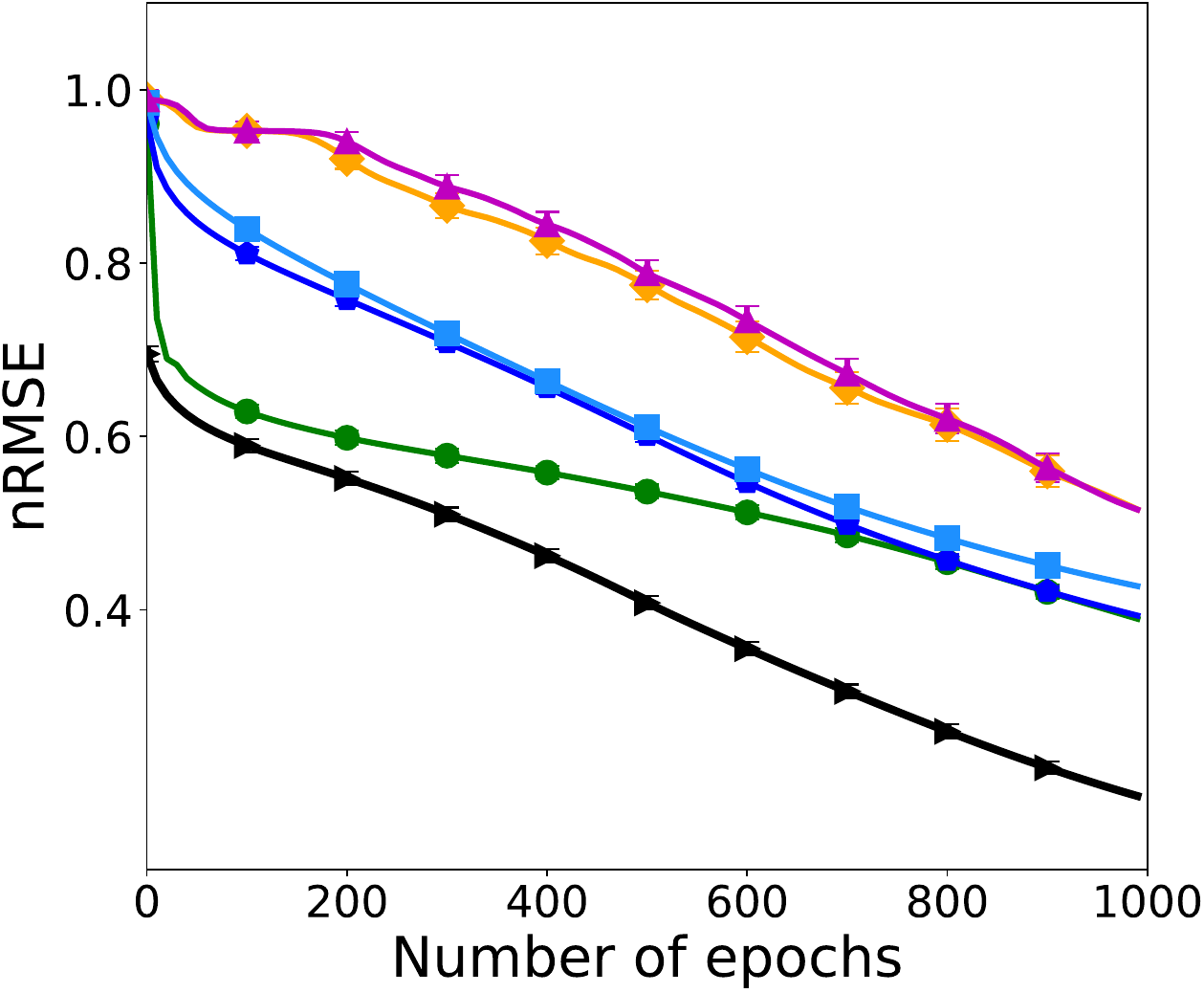}
		\caption{\small \textit{Alpine}: 50shot-50val}
	\end{subfigure}
		&
		\begin{subfigure}[t]{0.25\textwidth}
			\centering
			\includegraphics[width=\textwidth]{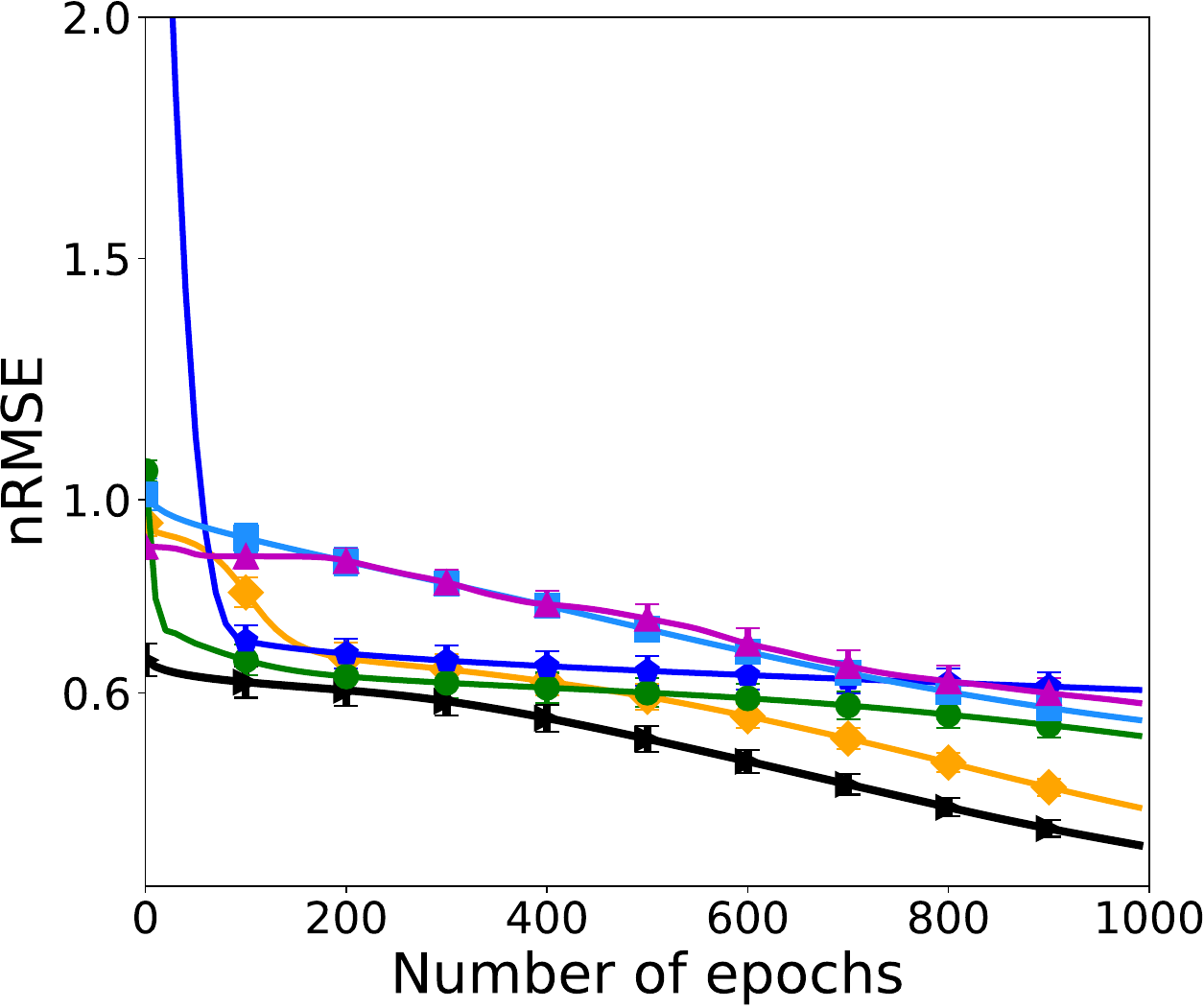}
			\caption{\small \textit{CosMixture}: 100shot-100val}
		\end{subfigure}  
		&
		\begin{subfigure}[t]{0.25\textwidth}
			\centering
			\includegraphics[width=\textwidth]{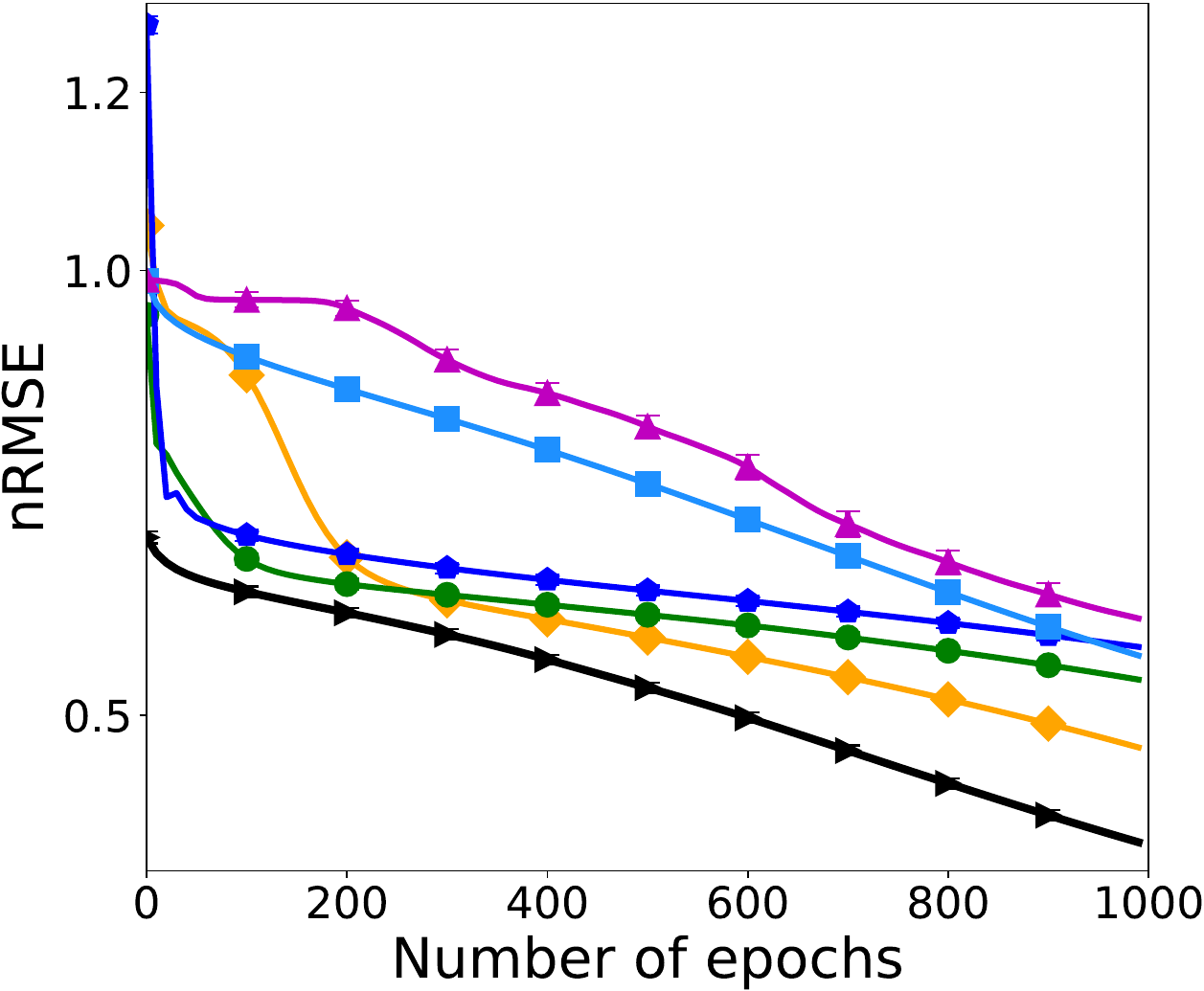}
			\caption{\small \textit{Alpine}: 100shot-100val}
		\end{subfigure} 
	\end{tabular}
	\vspace{-0.1in}
	\caption{\small Prediction error of the neural network in learning \textit{CosMixture} and \textit{Alpine} function families, starting from the initialization provided by different meta-learning approaches.  50shot-50val means 50 examples were used for meta-training and another 50 examples for meta-validation. 100shot-100val means both the meta-training and meta-validation used  100 examples. The results were averaged over 100 test tasks. } 	
	\label{fig:synthetic}
\end{figure*}
}

\noindent \textbf{Competing Methods.} To examine the effectiveness of our method \ours, \cmt{in computing the gradient in optimization based meta-learning,} we tested the following MAML based approaches for an \textit{apples-to-apples} comparison: (1) the original \maml~\citep{finn2017model}, (2) First-order \maml (\fomaml)~\citep{finn2017model}, which ignores the Jacobian in the gradient computation and uses the gradient \cmt{of the meta-validation loss} w.r.t. the trained parameters to update the initialization, (3) \rap~\citep{nichol2018first}, which subtracts the gradient w.r.t. the trained parameter by the current initialization as the updating direction, (4) Implicit \maml (\imaml)~\citep{rajeswaran2019meta}, which introduces a proximal regularizer in the meta-training loss, and uses conjugate gradient to compute the gradient w.r.t. the initialization. 

All the methods were implemented with PyTorch~\citep{paszke2019pytorch}. For \maml, we used a high-quality open source implementation ({\url{https://github.com/dragen1860/MAML-Pytorch}}); for \imaml, we used the  implementation of the original authors ({\url{https://github.com/aravindr93/imaml_dev}}). For our approach \ours, we used the Torchdiffeq library ({\url{https://github.com/rtqichen/torchdiffeq}}) to accomplish ODE solving with RK45. In the inner optimization, all the competing methods used the standard gradient descent (GD) with step size $\alpha = 0.01$. For \imaml, the strength of the proximal regularizer was chosen as $\lambda=1$ \cmt{\footnote{We also tested with other choices, \eg $0.5$ or $2$ as used in the original paper~\citep{rajeswaran2019meta}. The performance was almost the same.}} and 5 CG steps were conducted for Newton-CG optimization.  For our method, we used the same step size (\ie $0.01$) to run modified Euler's method~\citep{ascher1998computer} for solving the adjoint ODE. 
In the outer optimization, all the methods used the ADAM algorithm~\citep{kingma2014adam}, and the learning rate was set to $10^{-3}$. Each time, a mini-batch of five tasks were sampled to conduct inner-optimization, and then update the initialization in the outer-level. We ran $5,000$ meta-epochs for each method.  For the 50shot-50val setting, we ran $200$ GD steps for \fomaml, \rap, \imaml, and for our method \ours, set $T=2$ (that corresponds to 200 GD steps with $\alpha=0.01$). For the 100shot-100val setting, we ran 500 GD steps for \fomaml, \rap, \imaml, and set $T=5$ for \ours accordingly. By contrast, \maml ran $20$ and $50$ GD steps, respectively. Note that \maml cannot run too many GD steps without exhausting computational memory (see Section \ref{sect:memory}). We also evaluated \maml with only one GD step (the most common choice) for both settings; we denote such results by \maml-1. \zsdc{At the adaptation stage (meta-test), we ran the same number of GD steps with the initialization learned by every method:  200 steps for 50shot-50val and 500 steps for 100shot-100val, with the same step size as in the meta training.}
We executed all the algorithms  on a Linux workstation with an NVIDIA GeForce RTX 3090 GPU card that includes $24$ GB of G6X memory. 

In Fig. \ref{fig:synthetic}b,c, e and f, we show that starting with the learned initialization of each method, how the prediction error of the NN model on the test tasks varies along with the increase of training epochs.   The prediction error for each task is computed as the normalized root-mean-square error (nRMSE). We averaged the nRMSE over the $100$ test tasks and report the standard deviation. As we can see, in all the cases, our approach, \ours, always finds the initialization that leads to the best learning progress and performance  --- the NN models exhibit smaller prediction error throughout the training, as compared with using the initialization from the competing methods. \maml-1 is in general worse than \maml; the discrepancy is particularly evident for learning \textit{Alpine} functions with the 100shot-100val setting (see Fig. \ref{fig:synthetic}f). It implies that only performing one step GD in the inner-optimization might not properly reflect the quality of the initialization in training. Although \fomaml and \rap can run many GD steps, their performance is often worse than \maml, especially \rap, which is nearly always inferior to \maml. Such relatively poor performance might be attributed to the use of incorrect gradient information to update the initialization in these approaches. \imaml performed the second best at the beginning, but it was often surpassed by \maml or \fomaml after considerable training epochs. This might be due to (1) the proximity regularizer in the meta-training was not used in the actual training, which introduces some inconsistency, and (2) the inner optimization (though with 200/500 GD steps) has yet to achieve the optimum, and so the obtained gradient w.r.t. the initialization is still inaccurate. \zsdc{Note that the nRMSE for 100shot-100val seems a bit higher than 50shot-50val at the early stage, which might because the former involves a double quantity of examples, hence needs more epochs to train better and exhibits  slower learning progress.} Together these results have demonstrated the advantage of our method in being able to accurately compute the gradient for long inner-optimization trajectories. 

\vspace{-0.05in}
\subsection{Memory Consumption and Running Time}\label{sect:memory}
\begin{figure}[!htb]
	\centering
	\includegraphics[width=0.45\textwidth]{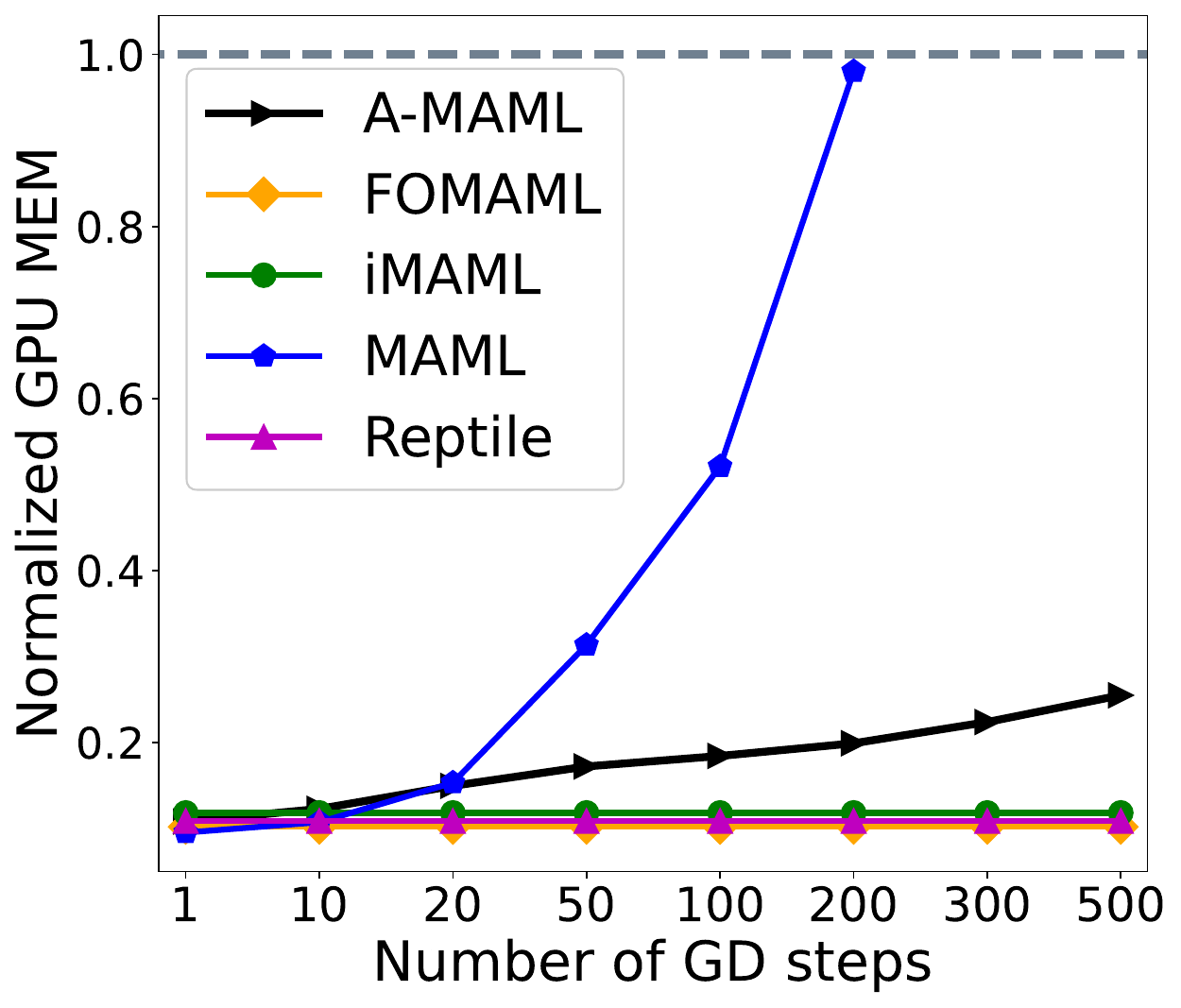}
	\caption{\small Normalized GPU usage in meta learning of \textit{CosMixutre} with 100shot-100validation. The dashed line indicates the capacity of available GPU memory.}
	\label{fig:mem-usage}
	\vspace{-0.1in}
\end{figure}

\begin{figure}[!htb]
	\centering
	\includegraphics[width=0.45\textwidth]{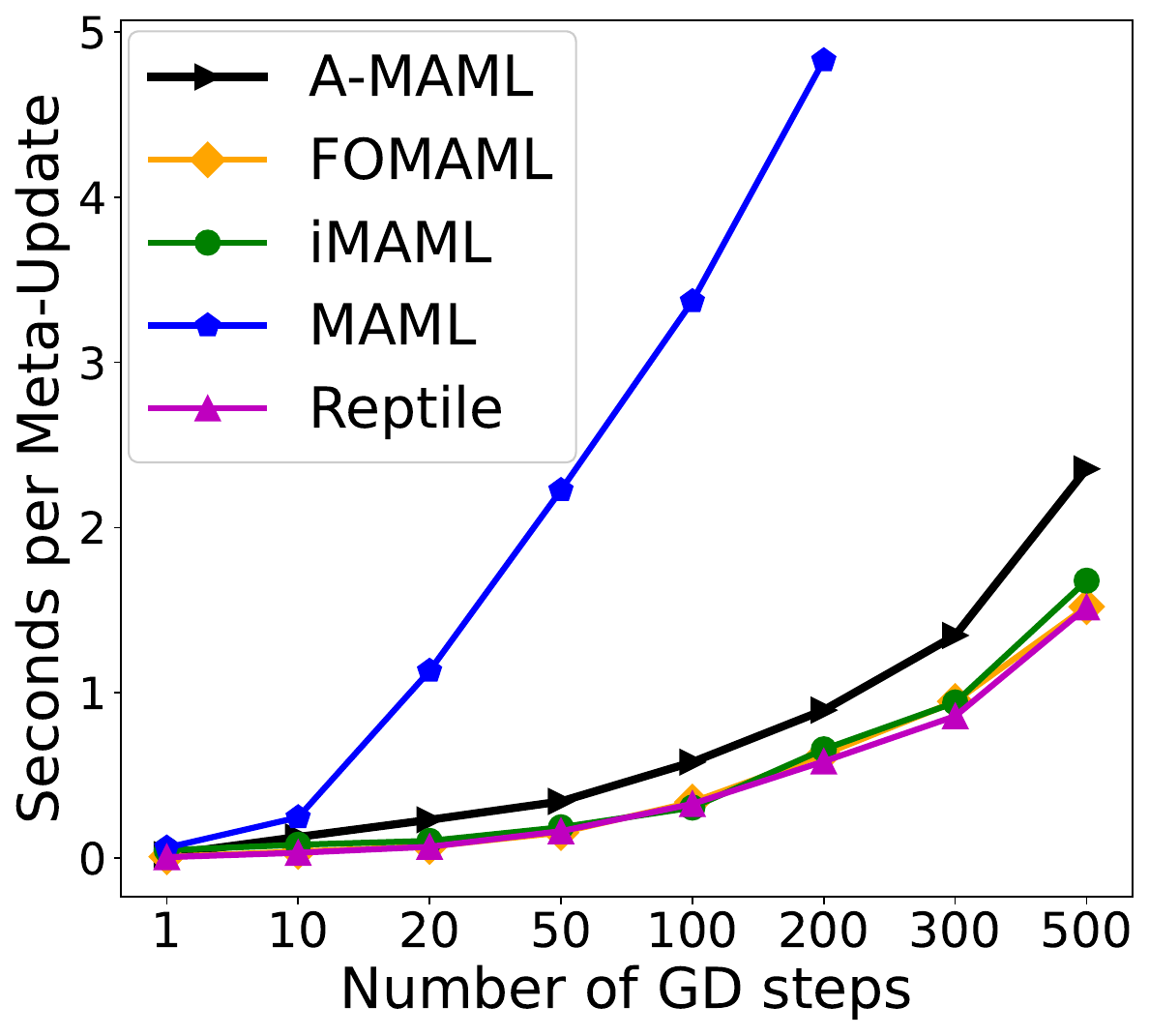}
	\caption{\small Running time of the inner gradient descent for \textit{CosMixutre}.}
	\label{fig:speed}
\end{figure}

\vspace{-0.1in}
Next we examined the efficiency of our method in terms of memory usage and computational speed. To this end, we tested the 100shot-100validation setting in the meta learning of \textit{CosMinxture} functions. We varied the number of inner GD steps (with the step size $\alpha = 0.01$) for the competing approaches and the corresponding time ranges $[0, T]$ for ODEs in \ours. The average memory usage and running time are reported in Fig. \ref{fig:mem-usage} and \ref{fig:speed}, respectively. 

\cmt{
\begin{figure}
	\centering
	\setlength{\tabcolsep}{0pt}
	\begin{tabular}[c]{c}
		\begin{subfigure}[b]{0.25\textwidth}
			\centering
			\includegraphics[width=\linewidth]{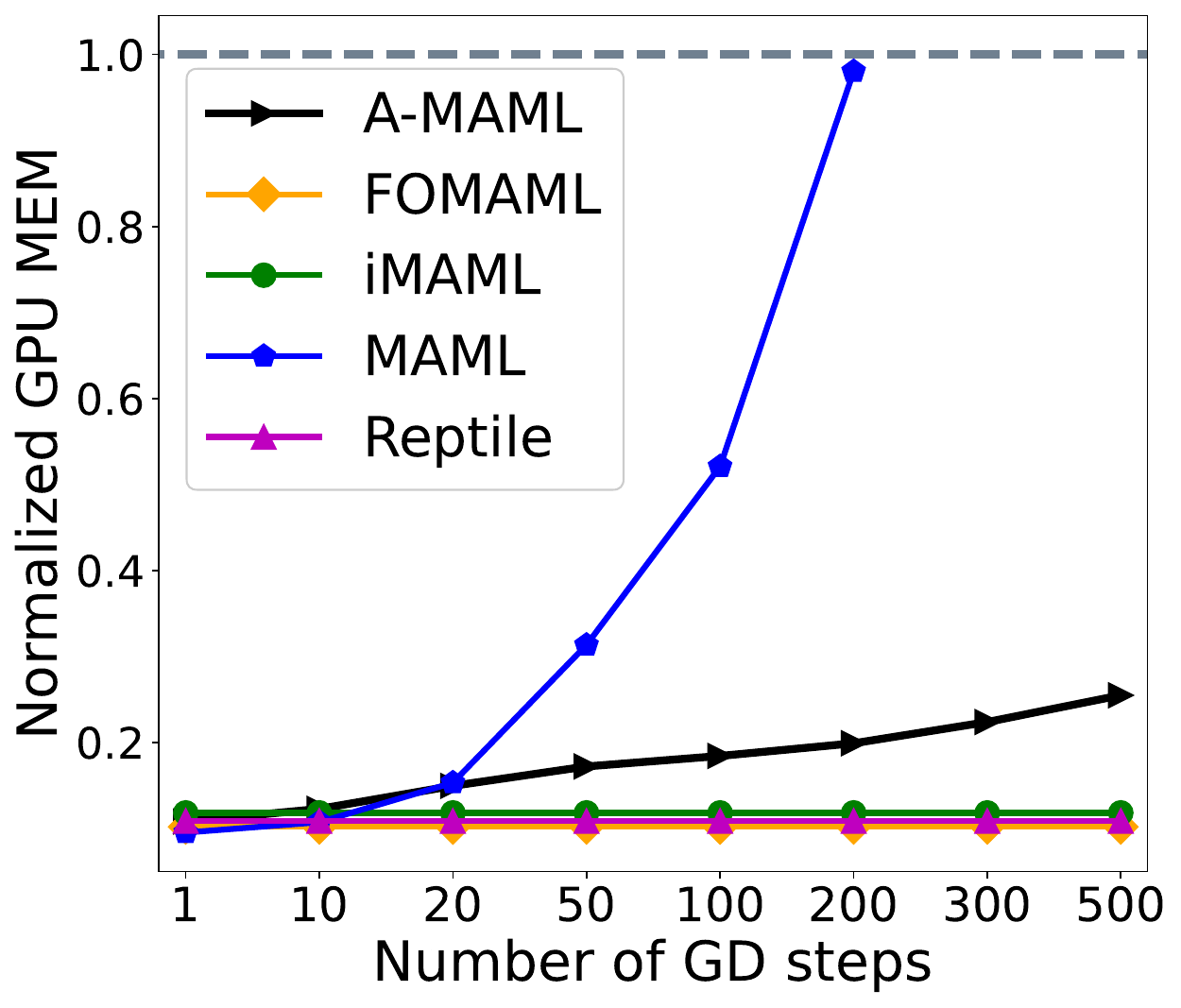}
			\caption{}
		\end{subfigure} \\
		\begin{subfigure}[b]{0.235\textwidth}
			\centering
			\includegraphics[width=\linewidth]{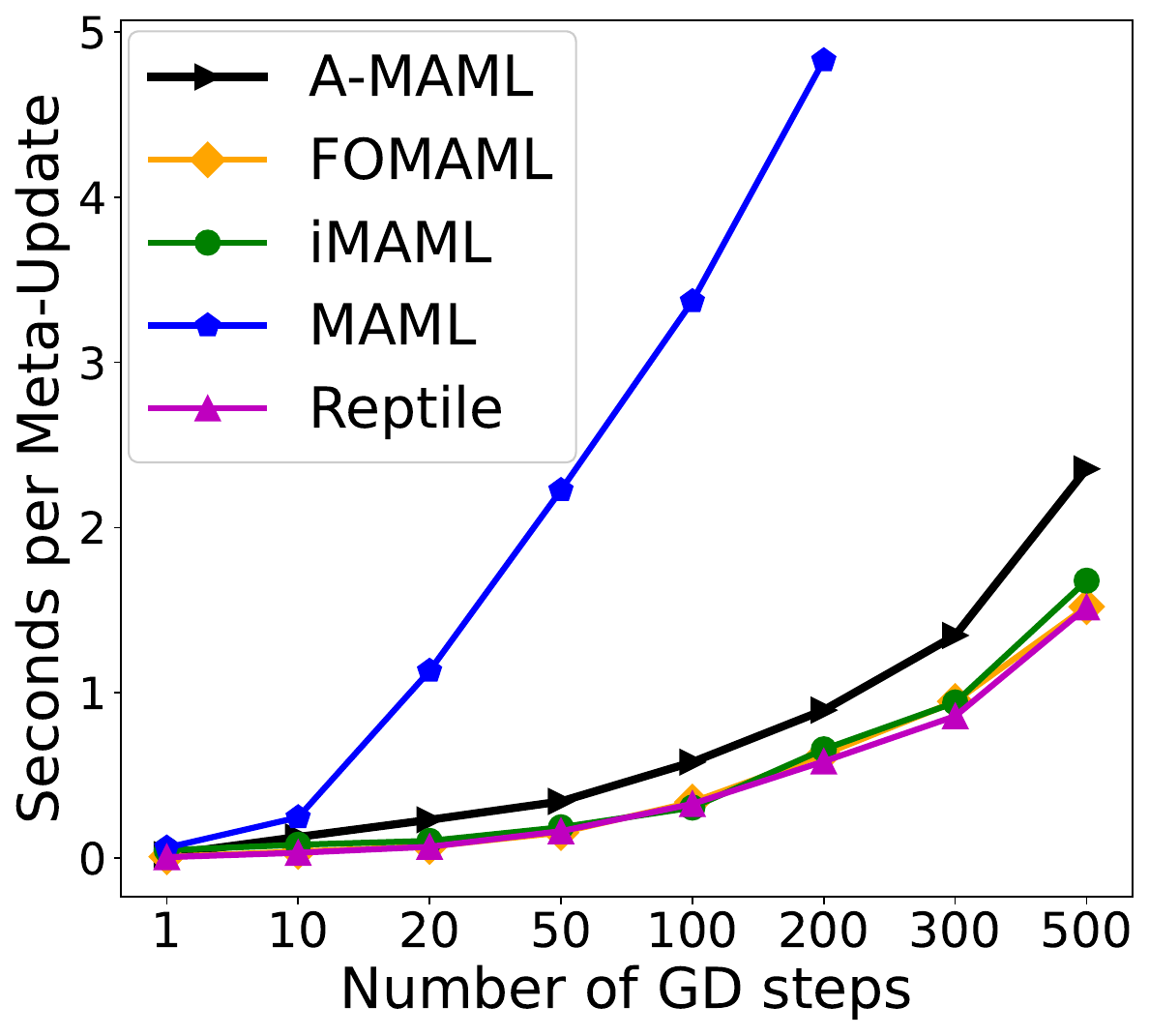}
			\caption{} 
		\end{subfigure} 
	\end{tabular}
	\caption{\small Normalized GPU usage (a) and running time (b) in meta learning of \textit{CosMixutre} with 100shot-100validation.The dashed line in (a) indicates the capacity of available GPU memory.}
	\vspace{-0.1in}
	\label{fig:efficiency}
\end{figure}
}

As shown in Fig. \ref{fig:mem-usage}, \maml always occupies the most memory. With the increase of GD steps, its memory consumption grows exponentially. When \maml runs 200 inner GD steps, the memory is completely exhausted.  The result shows the creation and expansion of the computation graphs is very costly. By contrast,  \ours can accurately compute the gradient in a much more economical way. \ours needs to track the states in the training trajectory to robustly solve the adjoint ODE so the memory usage also grows with the number of GD steps, but this growth is much slower (linear) and more affordable than \maml. \ours effortlessly supports 500 steps with less than 25\% memory usage.

 Fig. \ref{fig:speed} shows that the running time of \ours (per update in the outer-optimization) is comparable to  \imaml, \fomaml and \rap, and much smaller than \maml. This shows that our method is computationally efficient. On the other hand, the running time of \maml indicates that growing the computation graph for more GD steps also incurs a dramatic increase in the computational cost. 

\begin{table*}[]
	\centering
	\small
	\resizebox{\columnwidth}{!}{%
\begin{tabular}{|l|l|l|l|l|l|l|}
	\hline
	\multirow{2}{*}{} & \multicolumn{2}{c|}{Jester-1}                                             & \multicolumn{2}{c|}{MovieLens100K}                                        & \multicolumn{2}{c|}{MovieLens1M}                                          \\ \cline{2-7} 
	& \multicolumn{1}{c|}{10shot-15val} & \multicolumn{1}{c|}{20shot-30val} & \multicolumn{1}{c|}{10shot-15val} & \multicolumn{1}{c|}{20shot-30val} & \multicolumn{1}{c|}{10shot-15val} & \multicolumn{1}{c|}{20shot-30val} \\ \hline
	A-MAML            & \textbf{0.074$\pm$0.005 }                    & \textbf{0.027$\pm$0.002  }                   & \textbf{0.053$\pm$0.005 }                    & \textbf{0.023$\pm$0.003 }                    & \textbf{0.094$\pm$0.008  }                   & \textbf{0.035$\pm$0.004 }                    \\ \hline
	iMAML             & 0.114$\pm$0.007                     & 0.050$\pm$0.003                     & 0.082$\pm$0.004                     & 0.033$\pm$0.002                     & 0.138$\pm$0.010                     & 0.052$\pm$0.004                     \\ \hline
	MAML              & 0.120$\pm$0.001                     & 0.036$\pm$0.000                     & 0.123$\pm$0.001                     & 0.050$\pm$0.003                     & 0.140$\pm$0.002                     & 0.059$\pm$0.001                     \\ \hline
	FOMAML            & 0.292$\pm$0.012                     & 0.115$\pm$0.004                     & 0.174$\pm$0.008                     & 0.068$\pm$0.004                     & 0.270$\pm$0.011                     & 0.104$\pm$0.006                     \\ \hline
	Reptile           & 0.270$\pm$0.012                     & 0.106$\pm$0.004                     & 0.166$\pm$0.008                     & 0.063$\pm$0.003                     & 0.266$\pm$0.011                     & 0.101$\pm$0.006                     \\ \hline
\end{tabular}
}
	\caption{Meta-test error (nRMSE) with 50 inner GD steps (\maml used 5 GD steps). The results were averaged over $100$ tasks.} \label{tb:res1}
	\vspace{-0.1in}
\end{table*}

\begin{table*}[]
	\centering
	\small
	\resizebox{\columnwidth}{!}{%
\begin{tabular}{|l|l|l|l|l|l|l|}
	\hline
	\multirow{2}{*}{} & \multicolumn{2}{c|}{Jester-1}                                         & \multicolumn{2}{c|}{MovieLens100K}                                    & \multicolumn{2}{c|}{MovieLens1M}                                      \\ \cline{2-7} 
	& \multicolumn{1}{c|}{10shot-15val} & \multicolumn{1}{c|}{20shot-30val} & \multicolumn{1}{c|}{10shot-15val} & \multicolumn{1}{c|}{20shot-30val} & \multicolumn{1}{c|}{10shot-15val} & \multicolumn{1}{c|}{20shot-30val} \\ \hline
	A-MAML            & \textbf{0.069$\pm$0.005 }                  & \textbf{0.044$\pm$0.003 }                  & \textbf{0.057$\pm$0.006 }                  & \textbf{0.021$\pm$0.002 }                  & \textbf{0.105$\pm$0.009   }                & \textbf{0.035$\pm$0.004   }                \\ \hline
	iMAML             & 0.190$\pm$0.010                   & 0.103$\pm$0.005                   & 0.168$\pm$0.007                   & 0.046$\pm$0.002                   & 0.130$\pm$0.007                   & 0.045$\pm$0.004                   \\ \hline
	MAML              & 0.154$\pm$0.001                   & 0.061$\pm$0.002                   & 0.123$\pm$0.001                   & 0.050$\pm$0.002                   & 0.197$\pm$0.002                   & 0.083$\pm$0.001                   \\ \hline
	FOMAML            & 0.273$\pm$0.012                   & 0.077$\pm$0.004                   & 0.191$\pm$0.007                   & 0.071$\pm$0.004                   & 0.395$\pm$0.010                   & 0.119$\pm$0.005                   \\ \hline
	Reptile           & 0.290$\pm$0.012                   & 0.100$\pm$0.004                   & 0.171$\pm$0.008                   & 0.066$\pm$0.004                   & 0.408$\pm$0.011                   & 0.128$\pm$0.006                   \\ \hline
\end{tabular}
}
	\caption{Meta-test error (nRMSE) with 100 inner GD steps (\maml used 10 GD steps). The results were averaged over $100$ tasks.} \label{tb:res2}
	\vspace{-0.15in}
\end{table*}

\vspace{-0.05in}
\subsection{Few-Shot Learning in Collaborative Filtering}
\vspace{-0.1in}
Third, we examined our approach in three real-world applications of collaborative filtering. To this end, we used the following datasets. (1) \textit{Jester-1}({\url{https://goldberg.berkeley.edu/jester-data/}})~\citep{goldberg2001eigentaste}, which are about joke ratings. There are $100$ jokes, rated by $24,983$ users. Each user has rated at least $36$ jokes. The ratings are between -10 and 10. (2) \textit{MovieLens-100K}  and (3) \textit{MovieLens-1M} ({\url{https://grouplens.org/datasets/movielens/}}), movie rating datasets, where the former includes 10K ratings from 1K users on 1.7K movies, and the latter one million movie ratings from 6K users on 4K movies. The ratings are ranged from 0 to 5. 
Following~\citep{denevi2020advantage,denevi2021conditional}, we considered predicting the ratings of a given user (on different jokes or movies) as one task. 
\begin{table}
	\small
	\begin{center}
		\begin{small}
			\begin{tabular}{lcc}
				\toprule
				Method & {Ominiglot} & {Mini-ImageNet}  \\
				\midrule
				MAML   						   & $ 95.8 \pm 0.3\%$ & $48.70 \pm 1.84 \%$    \\
				FOMAML 					    & $ 89.4 \pm 0.5\% $  & $48.07 \pm 1.75 \%$  \\
				Reptile   					     &  $ 89.43 \pm 0.14\%$  & $\mathbf{49.97 \pm 0.32 \%}$ \\
				iMAML-GD 			       & $ 94.46 \pm 0.42\%$ & $48.96 \pm 1.84 \%$ \\
				iMAML-HF    			   &  $96.18 \pm 0.36\%$ & $49.30 \pm 1.88 \%$  \\
				A-MAML($T=0.5$)   & $96.36\pm 0.39 \%$   & $49.43 \pm 1.64\%$  \\
				A-MAML($T=1.0$)    & $\mathbf{96.79 \pm 0.34\%}$   & $49.47 \pm 1.77\%$  \\
				\bottomrule
			\end{tabular}
			\caption{\small Meta-test accuracy for 20-way 1-shot on \textit{Omniglot} and 5-way 1-shot on \textit{Mini-ImageNet}.} \label{tb:image}
		\end{small}
	\end{center}
\end{table}
Different users correspond to different tasks. For each user, we learned a neural network (NN) to predict the rating on a specific joke or movie. The input to the NN is the one-hot encoding of the joke or movie. The NN has two hidden layers, and each layer includes 40 neurons with Tanh activation. We conducted meta learning on each dataset to estimate a good initialization for the corresponding rate prediction model. To prevent scarcity of the task data points, we selected the most frequently rated 100 movies in \textit{MovieLens-100K} and \textit{MovieLens-1M}, and only considered users who had rated at least 20 of them. This gives 489 and 4,985 tasks  on \textit{MovieLens-100K} and \textit{MovieLens-1M}, respectively. For \textit{Jester-1}, we used all $24,983$ tasks. For each dataset, we sampled $100$ tasks for testing and used the remaining tasks  for meta learning. We examined two few-shot settings: 15shot-20val,  where 15 examples were used in meta-training and 20 examples in meta-validation, and 20shot-30val where 20 examples were used in meta-training and 30 example in meta-validation. During the meta learning, when the data points of a sampled task are less than the required meta training and validation set size, we re-sample a new task. 
At the test stage, the training for each task used the same number of examples for few-shot learning and the remaining were used for evaluation. 

For all the methods, the step size of the inner training was set to $\alpha = 0.01$, and a mini-batch of 5 tasks were sampled each time to conduct the inner training. We tested two choices of GD steps. First, we performed $50$ GD steps for \imaml, \fomaml and \rap, and set $T=0.5$ for \ours to solve the forward and adjoint ODEs (corresponding to $50$ steps). Second, we performed $100$ GD steps for \imaml, \fomaml and \rap, and accordingly set $T=1.0$ for \ours.  In each case, we ran \maml with one tenth of the corresponding steps, \ie $5$ and $10$ steps respectively.  
In the outer-level, all the methods used ADAM optimization with learning rate $10^{-3}$. We ran 5000 meta epochs for each method. We computed the average nRMSE and its standard deviation of using the initialization estimated by each method for training and then testing on new tasks. 

As shown in Tables \ref{tb:res1} and \ref{tb:res2}, \ours achieved the best performance in all the cases --- the learned initializations always result in the smallest test error after training ($p<0.05$), as compared with the competing methods. Consistent with the results in synthetic data (Sec. \ref{sect:syn}), \fomaml and \rap are still worse than \maml, implying that their updates with inaccurate gradient information do not help improve the performance in these collaborative filtering applications. The results further confirm the advantage of the proposed method \ours. 


\vspace{-0.05in}
\subsection{Few-Shot Learning in Images Classification}
\vspace{-0.05in}
Finally, we evaluated \ours on popular benchmark datasets in few-shot image classification tasks,  \textit{Mini-ImageNet} and \textit{Omniglot}. We followed the standard training and evaluation protocol as in \imaml paper and the prior works~\citep{santoro2016meta,vinyals2016matching,finn2017model}, including data splits, NN architecture, \etc We tested  5-way 1-shot learning on \textit{Mini-ImageNet} and 20-way 1-shot in \textit{Omniglot}, \textit{because these two settings are more challenging to all the methods.} We ran \ours with two settings, $T=0.1$ and $T=0.3$. \cmt{We solved the adjoint ODE backward with step size $0.01$.} During the adaptation stage, we ran the same number of GD steps with \imaml. 
The results are reported in Table \ref{tb:image}. \cmt{Note that iMAML-HF represents Hessian free CG (used in our other experiments) and iMAML-GD only CG.   }As we can see, with longer trajectory length, \ie $T=0.3$, our method gave the best performance on \textit{Omiglot} and the second best on \textit{Mini-ImageNet}. With shorter length ($T=0.1$), the performance decreases, but is comparable to or better than the competing methods. This is reasonable and again shows the advantage of being able to carry out longer trajectories during the meta training. 


\vspace{-0.05in}
\section{Conclusion}
\vspace{-0.1in}
We have presented  \ours, a novel meta learning approach of model initializations. We view the inner-optimization as solving a forward ODE, and use the adjoint method to compute the gradient of the meta-loss w.r.t. the initialization in an efficient and accurate way. 
We plan to extend our work to conditional meta learning~\citep{denevi2021conditional,wang2020structured} so as to further leverage side information to estimate task-specific initializations.

\newpage

\appendix

\section{Trade-off Analysis of Backward Solving} \label{sect:trade-off}
In this section, we examined the trade-off between the number of stored intermediate states and the accuracy of meta-gradient computation. To this end, we first considered a nonlinear ODE system, for which the gradient w.r.t the initial state has a closed form: 
\begin{align}
	\frac{\d y}{\d t} = -2x^{3} - 2ty.
\end{align}

The solution of the ODE is  
\begin{align}
	y(t) = 1 - t^2 + ce^{-t^2}
\end{align}
where $c$ is an arbitrary number and determined by the initial state, $c = y(0) - 1$.  We then define a synthetic objective function,  
\begin{align}
	\Lcal(y(t)) = (y(t)-3)^6.
\end{align}
Via the chain rule, we can obtain the gradient of the objective w.r.t  to the initial state $y(0)$, \ie the meta gradient,
\begin{align}
	\frac{\d \Lcal}{\d y(0)} = \frac{\d\Lcal}{\d y(t)}\cdot\frac{\d y(t)}{\d y(0)} = 6(y(t)-3)^5 e^{-t^2}.
\end{align}
We then examined the relative $L_2$ error of the meta gradient calculation by our method, with different $T$'s and numbers of intermediate states tracked in the back-solving process. For comparison, we tested the automatic differentiation (Autodiff) method based on computational graphs. To be fair, we used the same number of states to set the step size in the forward solving with the modified Euler method, and then applied Autodiff to compute the meta gradient. We repeated the experiment for $20$ times, and each time we used a random initial state. The results are reported in Table \ref{tab:syn-trade-off}.  Note that our method uses DPORI5 (Runge-Kutta of order 5 of Dormand-Prince-Shampine) for the forward solving and the modified Euler for the backward solving. 
We can see that the accuracy of our method is better than or comparable to Autodiff in all the cases. 
Tracking more intermediate states consistently improves the accuracy, yet bringing more memory consumption and computational cost. Hence, it enables us to select the cost and accuracy trade-off. 

\begin{table*}[ht]
	\footnotesize
	\begin{subtable}{0.8\textwidth}
		\begin{tabular}{|c|cc|cc|}
			\hline
			\multirow{2}{*}{Number of Intermediate States} & \multicolumn{2}{c|}{$T=0.1$}                       & \multicolumn{2}{c|}{$T=0.5$}                       \\ \cline{2-5}
			& \multicolumn{1}{c|}{Autodiff}      & Adjoint       & \multicolumn{1}{c|}{Autodiff}      & Adjoint       \\ \hline
			10   & \multicolumn{1}{c|}{6.24e-7 $\pm$ 8.87e-8} & 1.02e-6 $\pm$ 9.21e-7 & \multicolumn{1}{c|}{3.34e-4 $\pm$ 1.06e-7} & 2.67e-4 $\pm$ 1.79e-4 \\ \hline
			20   & \multicolumn{1}{c|}{2.01e-7 $\pm$ 1.08e-7} & 6.34e-7 $\pm$ 7.28e-7 & \multicolumn{1}{c|}{7.34e-5 $\pm$ 1.61e-7} & 6.17e-5 $\pm$ 4.07e-5\\ \hline
			50   & \multicolumn{1}{c|}{1.48e-7 $\pm$ 1.20e-7} & 6.88e-7 $\pm$ 7.49e-7 & \multicolumn{1}{c|}{1.08e-5 $\pm$ 2.49e-7} & 9.35e-6 $\pm$ 5.94e-6 \\ \hline
			100  & \multicolumn{1}{c|}{6.75e-7 $\pm$ 2.00e-7} & 5.87e-7 $\pm$ 5.31e-7 & \multicolumn{1}{c|}{2.64e-6 $\pm$ 2.59e-7} & 2.17e-6 $\pm$ 1.33e-6 \\ \hline
			200  & \multicolumn{1}{c|}{7.32e-7 $\pm$ 7.61e-7} & 4.49e-7 $\pm$ 3.62e-7 & \multicolumn{1}{c|}{9.18e-7 $\pm$ 4.13e-7} & 6.73e-7 $\pm$ 4.93e-7 \\ \hline
			500  & \multicolumn{1}{c|}{7.69e-7 $\pm$ 1.13e-6} & 3.01e-7 $\pm$ 1.99e-7 & \multicolumn{1}{c|}{1.12e-6 $\pm$ 6.59e-7} & 7.36e-7 $\pm$ 7.44e-7 \\ \hline
			1000 & \multicolumn{1}{c|}{7.09e-7 $\pm$ 7.82e-7} & 1.45e-7 $\pm$ 8.58e-8 & \multicolumn{1}{c|}{4.55e-6 $\pm$ 1.94e-6} & 6.94e-7 $\pm$ 4.73e-7 \\ \hline
		\end{tabular}
		\caption{}
		\label{tab:my-table}
	\end{subtable}
	\begin{subtable}{0.8\textwidth}
		\begin{tabular}{|c|cc|cc|}
			\hline
			\multirow{2}{*}{Number of Intermediate States} 	& \multicolumn{2}{c|}{$T=1.0$}                       & \multicolumn{2}{c|}{$T=2.0$}                       \\ \cline{2-5}
			& \multicolumn{1}{c|}{Autodiff}      & Adjoint       & \multicolumn{1}{c|}{Autodiff}      & Adjoint       \\ \hline
			10   & \multicolumn{1}{c|}{2.51e-3 $\pm$ 9.23e-4} & 2.48e-3 $\pm$ 7.28e-8 & \multicolumn{1}{c|}{4.74e-1 $\pm$ 4.13e-3} & 2.92e-1 $\pm$ 1.41e-7 \\ \hline
			20   & \multicolumn{1}{c|}{5.82e-4 $\pm$ 1.89e-4} & 5.04e-4 $\pm$ 1.64e-7 & \multicolumn{1}{c|}{7.33e-2 $\pm$ 4.91e-4} & 4.77e-2 $\pm$ 2.00e-7 \\ \hline
			50   & \multicolumn{1}{c|}{8.90e-5 $\pm$ 2.71e-5} & 7.17e-5 $\pm$ 2.14e-7 & \multicolumn{1}{c|}{9.28e-3 $\pm$ 5.93e-5} & 6.10e-3 $\pm$ 2.51e-7 \\ \hline
			100  & \multicolumn{1}{c|}{2.19e-5 $\pm$ 6.69e-6} & 1.73e-5 $\pm$ 3.40e-7 & \multicolumn{1}{c|}{2.16e-3 $\pm$ 1.38e-5} & 1.42e-3 $\pm$ 2.88e-7 \\ \hline
			200  & \multicolumn{1}{c|}{5.54e-6 $\pm$ 1.92e-6} & 4.34e-6 $\pm$ 4.82e-7 & \multicolumn{1}{c|}{5.23e-4 $\pm$ 3.76e-6} & 3.44e-4 $\pm$ 5.31e-7 \\ \hline
			500  & \multicolumn{1}{c|}{1.16e-6 $\pm$ 1.02e-6} & 9.91e-7 $\pm$ 5.42e-7 & \multicolumn{1}{c|}{8.14e-5 $\pm$ 1.27e-6} & 5.39e-5 $\pm$ 6.10e-7 \\ \hline
			1000 & \multicolumn{1}{c|}{1.51e-6 $\pm$ 1.51e-6} & 9.96e-7 $\pm$ 7.64e-7 & \multicolumn{1}{c|}{2.01e-5 $\pm$ 1.34e-6} & 1.32e-5 $\pm$ 1.05e-6 \\ \hline
		\end{tabular}
		\caption{}
		\label{tab:my-table2}
	\end{subtable}
	\caption{The relative $L_2$ error of meta-gradient computation. The results were averaged over $20$ random initializations. } \label{tab:syn-trade-off}
\end{table*}

Next, we examined the trade-off in the 2D regression problem, \textit{CosMixture} (see Sec. \ref{sect:syn}). In this problem, we do not have the ground-truth of the meta gradient. To evaluate the trade-off, we used the meta-gradient computed with $1,000$ intermediate states by our method as a reference. We then examined how the computed gradients using different numbers of states are close to the reference. We varied $T$ from $\{0.1, 0.3, 0.5\}$, and the number of intermediate steps from $\{10, 20, 50, 100, 200, 500\}$. We computed the relative $L_2$ error w.r.t the reference gradient. We tested on $20$ random initializations. The results are reported in Table \ref{tab:cosmix-trade-off}. As we can see, when only using $100$ or $200$ states, the computed meta gradient has already been very close to the one computed with 1,000 states. It implies that the gain of the accuracy is minor after a certain number of intermediate states. Hence, it is unnecessary to use too many states, and we can use much fewer to improve both the memory and computation efficiency. 

\begin{table*}[ht]
	\centering
	\begin{tabular}{|c|c|c|c|}
		\hline
		Number of Intermediate States	& $T=0.1$       & $T=0.3$       & $T=0.5$       \\ \hline
		10  & 7.62e-4 $\pm$ 2.04e-4 & 4.82e-3 $\pm$ 9.07e-4 & 1.17e-2 $\pm$ 2.17e-3 \\ \hline
		20  & 3.55e-4 $\pm$ 9.49e-5 & 2.26e-3 $\pm$ 4.21e-4 & 5.46e-3 $\pm$ 1.01e-3 \\ \hline
		50  & 1.33e-4 $\pm$ 3.55e-5 & 8.63e-4 $\pm$ 1.61e-4 & 2.09e-3 $\pm$ 3.85e-4 \\ \hline
		100 & 6.24e-5 $\pm$ 1.67e-5 & 4.19e-4 $\pm$ 7.79e-5 & 1.03e-3 $\pm$  1.88e-4 \\ \hline
		200 & 2.76e-5 $\pm$ 7.34e-6 & 2.01e-4 $\pm$ 3.74e-5 & 4.99e-4 $\pm$ 9.16e-5 \\ \hline
		500 & 6.87e-6 $\pm$ 1.83e-6 & 7.16e-5 $\pm$ 1.33e-5 & 1.87e-4 $\pm$ 3.42e-5 \\ \hline
	\end{tabular}
	\caption{The relative $L_2$ error w.r.t the gradient computed with 1K intermediate states on the \textit{CosMixture} problem. The results were averaged from 20 random initializations.}
	\label{tab:cosmix-trade-off}
\end{table*}

\bibliography{AMAML}

\begin{thebibliography}{53}
\providecommand{\natexlab}[1]{#1}
\providecommand{\url}[1]{\texttt{#1}}
\expandafter\ifx\csname urlstyle\endcsname\relax
  \providecommand{\doi}[1]{doi: #1}\else
  \providecommand{\doi}{doi: \begingroup \urlstyle{rm}\Url}\fi

\bibitem[Allen et~al.(2019)Allen, Shelhamer, Shin, and
  Tenenbaum]{allen2019infinite}
Kelsey Allen, Evan Shelhamer, Hanul Shin, and Joshua Tenenbaum.
\newblock Infinite mixture prototypes for few-shot learning.
\newblock In \emph{International Conference on Machine Learning}, pages
  232--241. PMLR, 2019.

\bibitem[Andrychowicz et~al.(2016)Andrychowicz, Denil, Gomez, Hoffman, Pfau,
  Schaul, Shillingford, and De~Freitas]{andrychowicz2016learning}
Marcin Andrychowicz, Misha Denil, Sergio Gomez, Matthew~W Hoffman, David Pfau,
  Tom Schaul, Brendan Shillingford, and Nando De~Freitas.
\newblock Learning to learn by gradient descent by gradient descent.
\newblock \emph{arXiv preprint arXiv:1606.04474}, 2016.

\bibitem[Ascher and Petzold(1998)]{ascher1998computer}
Uri~M Ascher and Linda~R Petzold.
\newblock \emph{Computer methods for ordinary differential equations and
  differential-algebraic equations}, volume~61.
\newblock Siam, 1998.

\bibitem[Baydin and Pearlmutter(2014)]{baydin2014automatic}
Atilim~Gunes Baydin and Barak~A Pearlmutter.
\newblock Automatic differentiation of algorithms for machine learning.
\newblock \emph{arXiv preprint arXiv:1404.7456}, 2014.

\bibitem[Bengio(2000)]{bengio2000gradient}
Yoshua Bengio.
\newblock Gradient-based optimization of hyperparameters.
\newblock \emph{Neural computation}, 12\penalty0 (8):\penalty0 1889--1900,
  2000.

\bibitem[Bertinetto et~al.(2018)Bertinetto, Henriques, Torr, and
  Vedaldi]{bertinetto2018meta}
Luca Bertinetto, Joao~F Henriques, Philip~HS Torr, and Andrea Vedaldi.
\newblock Meta-learning with differentiable closed-form solvers.
\newblock \emph{arXiv preprint arXiv:1805.08136}, 2018.

\bibitem[Chen et~al.(2018)Chen, Rubanova, Bettencourt, and
  Duvenaud]{chen2018neural}
Ricky~TQ Chen, Yulia Rubanova, Jesse Bettencourt, and David~K Duvenaud.
\newblock Neural ordinary differential equations.
\newblock \emph{Advances in neural information processing systems}, 31, 2018.

\bibitem[Denevi et~al.(2020)Denevi, Pontil, and Ciliberto]{denevi2020advantage}
Giulia Denevi, Massimiliano Pontil, and Carlo Ciliberto.
\newblock The advantage of conditional meta-learning for biased regularization
  and fine tuning.
\newblock \emph{Advances in Neural Information Processing Systems}, 33, 2020.

\bibitem[Denevi et~al.(2021)Denevi, Pontil, and
  Ciliberto]{denevi2021conditional}
Giulia Denevi, Massimiliano Pontil, and Carlo Ciliberto.
\newblock Conditional meta-learning of linear representations.
\newblock \emph{arXiv preprint arXiv:2103.16277}, 2021.

\bibitem[Domke(2012)]{domke2012generic}
Justin Domke.
\newblock Generic methods for optimization-based modeling.
\newblock In \emph{Artificial Intelligence and Statistics}, pages 318--326.
  PMLR, 2012.

\bibitem[Dormand and Prince(1980)]{dormand1980family}
John~R Dormand and Peter~J Prince.
\newblock A family of embedded runge-kutta formulae.
\newblock \emph{Journal of computational and applied mathematics}, 6\penalty0
  (1):\penalty0 19--26, 1980.

\bibitem[Duan et~al.(2016)Duan, Schulman, Chen, Bartlett, Sutskever, and
  Abbeel]{duan2016rl}
Yan Duan, John Schulman, Xi~Chen, Peter~L Bartlett, Ilya Sutskever, and Pieter
  Abbeel.
\newblock Rl2: Fast reinforcement learning via slow reinforcement learning.
\newblock \emph{arXiv preprint arXiv:1611.02779}, 2016.

\bibitem[Eichmeir et~al.(2021)Eichmeir, Lau{\ss}, Oberpeilsteiner, Nachbagauer,
  and Steiner]{eichmeir2021adjoint}
Philipp Eichmeir, Thomas Lau{\ss}, Stefan Oberpeilsteiner, Karin Nachbagauer,
  and Wolfgang Steiner.
\newblock The adjoint method for time-optimal control problems.
\newblock \emph{Journal of Computational and Nonlinear Dynamics}, 16\penalty0
  (2), 2021.

\bibitem[Finn(2018)]{finn2018learning}
Chelsea Finn.
\newblock \emph{Learning to learn with gradients}.
\newblock PhD thesis, UC Berkeley, 2018.

\bibitem[Finn et~al.(2017)Finn, Abbeel, and Levine]{finn2017model}
Chelsea Finn, Pieter Abbeel, and Sergey Levine.
\newblock Model-agnostic meta-learning for fast adaptation of deep networks.
\newblock In \emph{International Conference on Machine Learning}, pages
  1126--1135. PMLR, 2017.

\bibitem[Finn et~al.(2018)Finn, Xu, and Levine]{finn2018probabilistic}
Chelsea Finn, Kelvin Xu, and Sergey Levine.
\newblock Probabilistic model-agnostic meta-learning.
\newblock \emph{arXiv preprint arXiv:1806.02817}, 2018.

\bibitem[Goldberg et~al.(2001)Goldberg, Roeder, Gupta, and
  Perkins]{goldberg2001eigentaste}
Ken Goldberg, Theresa Roeder, Dhruv Gupta, and Chris Perkins.
\newblock Eigentaste: A constant time collaborative filtering algorithm.
\newblock \emph{information retrieval}, 4\penalty0 (2):\penalty0 133--151,
  2001.

\bibitem[Grant et~al.(2018)Grant, Finn, Levine, Darrell, and
  Griffiths]{grant2018recasting}
Erin Grant, Chelsea Finn, Sergey Levine, Trevor Darrell, and Thomas Griffiths.
\newblock Recasting gradient-based meta-learning as hierarchical {B}ayes.
\newblock In \emph{6th International Conference on Learning Representations,
  ICLR 2018}, 2018.

\bibitem[Harrison et~al.(2018)Harrison, Sharma, and Pavone]{harrison2018meta}
James Harrison, Apoorva Sharma, and Marco Pavone.
\newblock Meta-learning priors for efficient online bayesian regression.
\newblock In \emph{International Workshop on the Algorithmic Foundations of
  Robotics}, pages 318--337. Springer, 2018.

\bibitem[Hochreiter et~al.(2001)Hochreiter, Younger, and
  Conwell]{hochreiter2001learning}
Sepp Hochreiter, A~Steven Younger, and Peter~R Conwell.
\newblock Learning to learn using gradient descent.
\newblock In \emph{International Conference on Artificial Neural Networks},
  pages 87--94. Springer, 2001.

\bibitem[Hospedales et~al.(2020)Hospedales, Antoniou, Micaelli, and
  Storkey]{hospedales2020meta}
Timothy Hospedales, Antreas Antoniou, Paul Micaelli, and Amos Storkey.
\newblock Meta-learning in neural networks: A survey.
\newblock \emph{arXiv preprint arXiv:2004.05439}, 2020.

\bibitem[Im et~al.(2019)Im, Jiang, and Verma]{im2019model}
Daniel~Jiwoong Im, Yibo Jiang, and Nakul Verma.
\newblock Model-agnostic meta-learning using runge-kutta methods.
\newblock \emph{arXiv preprint arXiv:1910.07368}, 2019.

\bibitem[Kingma and Ba(2014)]{kingma2014adam}
Diederik~P Kingma and Jimmy Ba.
\newblock Adam: A method for stochastic optimization.
\newblock \emph{arXiv preprint arXiv:1412.6980}, 2014.

\bibitem[Koch et~al.(2015)Koch, Zemel, and Salakhutdinov]{koch2015siamese}
Gregory Koch, Richard Zemel, and Ruslan Salakhutdinov.
\newblock Siamese neural networks for one-shot image recognition.
\newblock In \emph{ICML deep learning workshop}, volume~2. Lille, 2015.

\bibitem[Lake et~al.(2011)Lake, Salakhutdinov, Gross, and
  Tenenbaum]{lake2011one}
Brenden Lake, Ruslan Salakhutdinov, Jason Gross, and Joshua Tenenbaum.
\newblock One shot learning of simple visual concepts.
\newblock In \emph{Proceedings of the annual meeting of the cognitive science
  society}, volume~33, 2011.

\bibitem[Lee et~al.(2019)Lee, Maji, Ravichandran, and Soatto]{lee2019meta}
Kwonjoon Lee, Subhransu Maji, Avinash Ravichandran, and Stefano Soatto.
\newblock Meta-learning with differentiable convex optimization.
\newblock In \emph{Proceedings of the IEEE/CVF Conference on Computer Vision
  and Pattern Recognition}, pages 10657--10665, 2019.

\bibitem[Li and Malik(2016)]{li2016learning}
Ke~Li and Jitendra Malik.
\newblock Learning to optimize.
\newblock \emph{arXiv preprint arXiv:1606.01885}, 2016.

\bibitem[Li et~al.(2017)Li, Zhou, Chen, and Li]{li2017meta}
Zhenguo Li, Fengwei Zhou, Fei Chen, and Hang Li.
\newblock Meta-sgd: Learning to learn quickly for few-shot learning.
\newblock \emph{arXiv preprint arXiv:1707.09835}, 2017.

\bibitem[Liu et~al.(2019)Liu, Socher, and Xiong]{liu2019taming}
Hao Liu, Richard Socher, and Caiming Xiong.
\newblock Taming maml: Efficient unbiased meta-reinforcement learning.
\newblock In \emph{International Conference on Machine Learning}, pages
  4061--4071. PMLR, 2019.

\bibitem[Mishra et~al.(2017)Mishra, Rohaninejad, Chen, and
  Abbeel]{mishra2017simple}
Nikhil Mishra, Mostafa Rohaninejad, Xi~Chen, and Pieter Abbeel.
\newblock A simple neural attentive meta-learner.
\newblock \emph{arXiv preprint arXiv:1707.03141}, 2017.

\bibitem[Munkhdalai and Yu(2017)]{munkhdalai2017meta}
Tsendsuren Munkhdalai and Hong Yu.
\newblock Meta networks.
\newblock In \emph{International Conference on Machine Learning}, pages
  2554--2563. PMLR, 2017.

\bibitem[Naik and Mammone(1992)]{naik1992meta}
Devang~K Naik and Richard~J Mammone.
\newblock Meta-neural networks that learn by learning.
\newblock In \emph{[Proceedings 1992] IJCNN International Joint Conference on
  Neural Networks}, volume~1, pages 437--442. IEEE, 1992.

\bibitem[Nichol et~al.(2018)Nichol, Achiam, and Schulman]{nichol2018first}
Alex Nichol, Joshua Achiam, and John Schulman.
\newblock On first-order meta-learning algorithms.
\newblock \emph{arXiv preprint arXiv:1803.02999}, 2018.

\bibitem[Oreshkin et~al.(2018)Oreshkin, Rodriguez, and
  Lacoste]{oreshkin2018tadam}
Boris~N Oreshkin, Pau Rodriguez, and Alexandre Lacoste.
\newblock Tadam: Task dependent adaptive metric for improved few-shot learning.
\newblock \emph{arXiv preprint arXiv:1805.10123}, 2018.

\bibitem[Paszke et~al.(2019)Paszke, Gross, Massa, Lerer, Bradbury, Chanan,
  Killeen, Lin, Gimelshein, Antiga, et~al.]{paszke2019pytorch}
Adam Paszke, Sam Gross, Francisco Massa, Adam Lerer, James Bradbury, Gregory
  Chanan, Trevor Killeen, Zeming Lin, Natalia Gimelshein, Luca Antiga, et~al.
\newblock Pytorch: An imperative style, high-performance deep learning library.
\newblock \emph{arXiv preprint arXiv:1912.01703}, 2019.

\bibitem[Pontryagin(1987)]{pontryagin1987mathematical}
Lev~Semenovich Pontryagin.
\newblock \emph{Mathematical theory of optimal processes}.
\newblock CRC press, 1987.

\bibitem[Rajeswaran et~al.(2019)Rajeswaran, Finn, Kakade, and
  Levine]{rajeswaran2019meta}
Aravind Rajeswaran, Chelsea Finn, Sham Kakade, and Sergey Levine.
\newblock Meta-learning with implicit gradients.
\newblock \emph{Advances in neural information processing systems}, 2019.

\bibitem[Ravi and Larochelle(2017)]{ravi2016optimization}
Sachin Ravi and Hugo Larochelle.
\newblock Optimization as a model for few-shot learning.
\newblock In \emph{In International Conference on Learning Representations
  (ICLR)}, 2017.

\bibitem[Rusu et~al.(2018)Rusu, Rao, Sygnowski, Vinyals, Pascanu, Osindero, and
  Hadsell]{rusu2018meta}
Andrei~A Rusu, Dushyant Rao, Jakub Sygnowski, Oriol Vinyals, Razvan Pascanu,
  Simon Osindero, and Raia Hadsell.
\newblock Meta-learning with latent embedding optimization.
\newblock \emph{arXiv preprint arXiv:1807.05960}, 2018.

\bibitem[Santoro et~al.(2016)Santoro, Bartunov, Botvinick, Wierstra, and
  Lillicrap]{santoro2016meta}
Adam Santoro, Sergey Bartunov, Matthew Botvinick, Daan Wierstra, and Timothy
  Lillicrap.
\newblock Meta-learning with memory-augmented neural networks.
\newblock In \emph{International conference on machine learning}, pages
  1842--1850. PMLR, 2016.

\bibitem[Schmidhuber(1987)]{schmidhuber1987evolutionary}
J{\"u}rgen Schmidhuber.
\newblock \emph{Evolutionary principles in self-referential learning, or on
  learning how to learn: the meta-meta-... hook}.
\newblock PhD thesis, Technische Universit{\"a}t M{\"u}nchen, 1987.

\bibitem[Snell et~al.(2017)Snell, Swersky, and Zemel]{snell2017prototypical}
Jake Snell, Kevin Swersky, and Richard Zemel.
\newblock Prototypical networks for few-shot learning.
\newblock \emph{Advances in neural information processing systems}, 30, 2017.

\bibitem[Song et~al.(2020)Song, Gao, Yang, Choromanski, Pacchiano, and
  Tang]{song2020maml}
Xingyou Song, Wenbo Gao, Yuxiang Yang, Krzysztof Choromanski, Aldo Pacchiano,
  and Yunhao Tang.
\newblock Es-maml: Simple hessian-free meta learning.
\newblock In \emph{ICLR}, 2020.

\bibitem[Sung et~al.(2018)Sung, Yang, Zhang, Xiang, Torr, and
  Hospedales]{sung2018learning}
Flood Sung, Yongxin Yang, Li~Zhang, Tao Xiang, Philip~HS Torr, and Timothy~M
  Hospedales.
\newblock Learning to compare: Relation network for few-shot learning.
\newblock In \emph{Proceedings of the IEEE conference on computer vision and
  pattern recognition}, pages 1199--1208, 2018.

\bibitem[Thrun and Pratt(2012)]{thrun2012learning}
Sebastian Thrun and Lorien Pratt.
\newblock \emph{Learning to learn}.
\newblock Springer Science \& Business Media, 2012.

\bibitem[Triantafillou et~al.(2019)Triantafillou, Zhu, Dumoulin, Lamblin, Evci,
  Xu, Goroshin, Gelada, Swersky, Manzagol, et~al.]{triantafillou2019meta}
Eleni Triantafillou, Tyler Zhu, Vincent Dumoulin, Pascal Lamblin, Utku Evci,
  Kelvin Xu, Ross Goroshin, Carles Gelada, Kevin Swersky, Pierre-Antoine
  Manzagol, et~al.
\newblock Meta-dataset: A dataset of datasets for learning to learn from few
  examples.
\newblock \emph{arXiv preprint arXiv:1903.03096}, 2019.

\bibitem[Vinyals et~al.(2016)Vinyals, Blundell, Lillicrap, Kavukcuoglu, and
  Wierstra]{vinyals2016matching}
Oriol Vinyals, Charles Blundell, Timothy Lillicrap, Koray Kavukcuoglu, and Daan
  Wierstra.
\newblock Matching networks for one shot learning.
\newblock \emph{arXiv preprint arXiv:1606.04080}, 2016.

\bibitem[Wang et~al.(2016)Wang, Kurth-Nelson, Tirumala, Soyer, Leibo, Munos,
  Blundell, Kumaran, and Botvinick]{wang2016learning}
Jane~X Wang, Zeb Kurth-Nelson, Dhruva Tirumala, Hubert Soyer, Joel~Z Leibo,
  Remi Munos, Charles Blundell, Dharshan Kumaran, and Matt Botvinick.
\newblock Learning to reinforcement learn.
\newblock \emph{arXiv preprint arXiv:1611.05763}, 2016.

\bibitem[Wang et~al.(2020)Wang, Demiris, and Ciliberto]{wang2020structured}
Ruohan Wang, Yiannis Demiris, and Carlo Ciliberto.
\newblock Structured prediction for conditional meta-learning.
\newblock \emph{Advances in Neural Information Processing Systems}, 33, 2020.

\bibitem[Xu et~al.(2021)Xu, Chen, and Karbasi]{xu2021meta}
Ruitu Xu, Lin Chen, and Amin Karbasi.
\newblock Meta learning in the continuous time limit.
\newblock In \emph{International Conference on Artificial Intelligence and
  Statistics}, pages 3052--3060. PMLR, 2021.

\bibitem[Yoon et~al.(2018)Yoon, Kim, Dia, Kim, Bengio, and
  Ahn]{yoon2018bayesian}
Jaesik Yoon, Taesup Kim, Ousmane Dia, Sungwoong Kim, Yoshua Bengio, and Sungjin
  Ahn.
\newblock Bayesian model-agnostic meta-learning.
\newblock In \emph{Proceedings of the 32nd International Conference on Neural
  Information Processing Systems}, pages 7343--7353, 2018.

\bibitem[Zhou et~al.(2018)Zhou, Wu, and Li]{zhou2018deep}
Fengwei Zhou, Bin Wu, and Zhenguo Li.
\newblock Deep meta-learning: Learning to learn in the concept space.
\newblock \emph{arXiv preprint arXiv:1802.03596}, 2018.

\bibitem[Zintgraf et~al.(2019)Zintgraf, Shiarli, Kurin, Hofmann, and
  Whiteson]{zintgraf2019fast}
Luisa Zintgraf, Kyriacos Shiarli, Vitaly Kurin, Katja Hofmann, and Shimon
  Whiteson.
\newblock Fast context adaptation via meta-learning.
\newblock In \emph{International Conference on Machine Learning}, pages
  7693--7702. PMLR, 2019.

\end{thebibliography}

\end{document}